\documentclass[review]{elsarticle}

\usepackage{lineno,xcolor}
\usepackage{amsmath,amssymb}
\usepackage{siunitx} 

\usepackage{enumitem}

\usepackage{wrapfig}
\usepackage{lscape}
\usepackage{rotating}
\usepackage{epstopdf}

\usepackage{tablefootnote}
\usepackage{soul}
\usepackage{subcaption} 
\usepackage{comment}

\usepackage[pagebackref=true,breaklinks=true,letterpaper=true,colorlinks,bookmarks=false]{hyperref}

\newcommand{\etal}{et al.}
\newcommand{\eg}{e.\,g.,}
\newcommand{\ie}{i.\,e.,}

\newcommand{\rw}[1]{\textcolor{blue}{#1}}

\newcommand{\xdownarrow}[1]{%
  {\left\downarrow\vbox to #1{}\right.\kern-\nulldelimiterspace}
}

\journal{ISPRS Journal of Photogrammetry and Remote Sensing}

\begin{document}

\begin{frontmatter}

\title{AMENet: Attentive Maps Encoder Network \\ for Trajectory Prediction}


\author[IKG]{Hao Cheng\fnref{fa1}}
\ead{hao.cheng@ikg.uni-hannover.de}

\author[TNT]{Wentong Liao\fnref{fa1}\corref{cor1}}
\ead{liao@tnt.uni-hannover.de}

\author[UT]{Michael Ying Yang\corref{cor1}}
\ead{michael.yang@utwente.nl}

\author[TNT]{Bodo Rosenhahn}
\ead{rosenhahn@tnt.uni-hannover.de}

\author[IKG]{Monika Sester}
\ead{monika.sester@ikg.uni-hannover.de}

\fntext[fa1]{Joint first author, arranged in alphabetical order.}
\cortext[cor1]{Corresponding authors}

\address[IKG]{Institute of Cartography
and Geoinformatics, Leibniz University Hannover, Germany}

\address[TNT]{Institute of Information
Processing, Leibniz University Hannover, Germany}

\address[UT]{Scene Understanding Group, Faculty of ITC, University of Twente, The Netherlands}

\begin{abstract}
Trajectory prediction is critical for applications of planning safe future movements and remains challenging even for the next few seconds in urban mixed traffic.
How an agent moves is affected by the various behaviors of its neighboring agents in different environments.
To predict movements, we propose an end-to-end generative model named \emph{Attentive Maps Encoder Network (AMENet)} that encodes the agent's motion and interaction information for accurate and realistic multi-path trajectory prediction.
A conditional variational auto-encoder module is trained to learn the latent space of possible future paths based on attentive dynamic maps for interaction modeling and then is used to predict multiple plausible future trajectories conditioned on the observed past trajectories.
The efficacy of AMENet is validated using two public trajectory prediction benchmarks \emph{Trajnet} and \emph{InD}.
\end{abstract}

\begin{keyword}
\texttt{Trajectory prediction\sep Generative model\sep Encoder}
\end{keyword}

\end{frontmatter}

\section{Introduction}
\label{sec:introduction}

\begin{figure}[t!]
    \centering
    \includegraphics[trim=0.6in 5.8in 2.6in 0.6in, width=\textwidth]{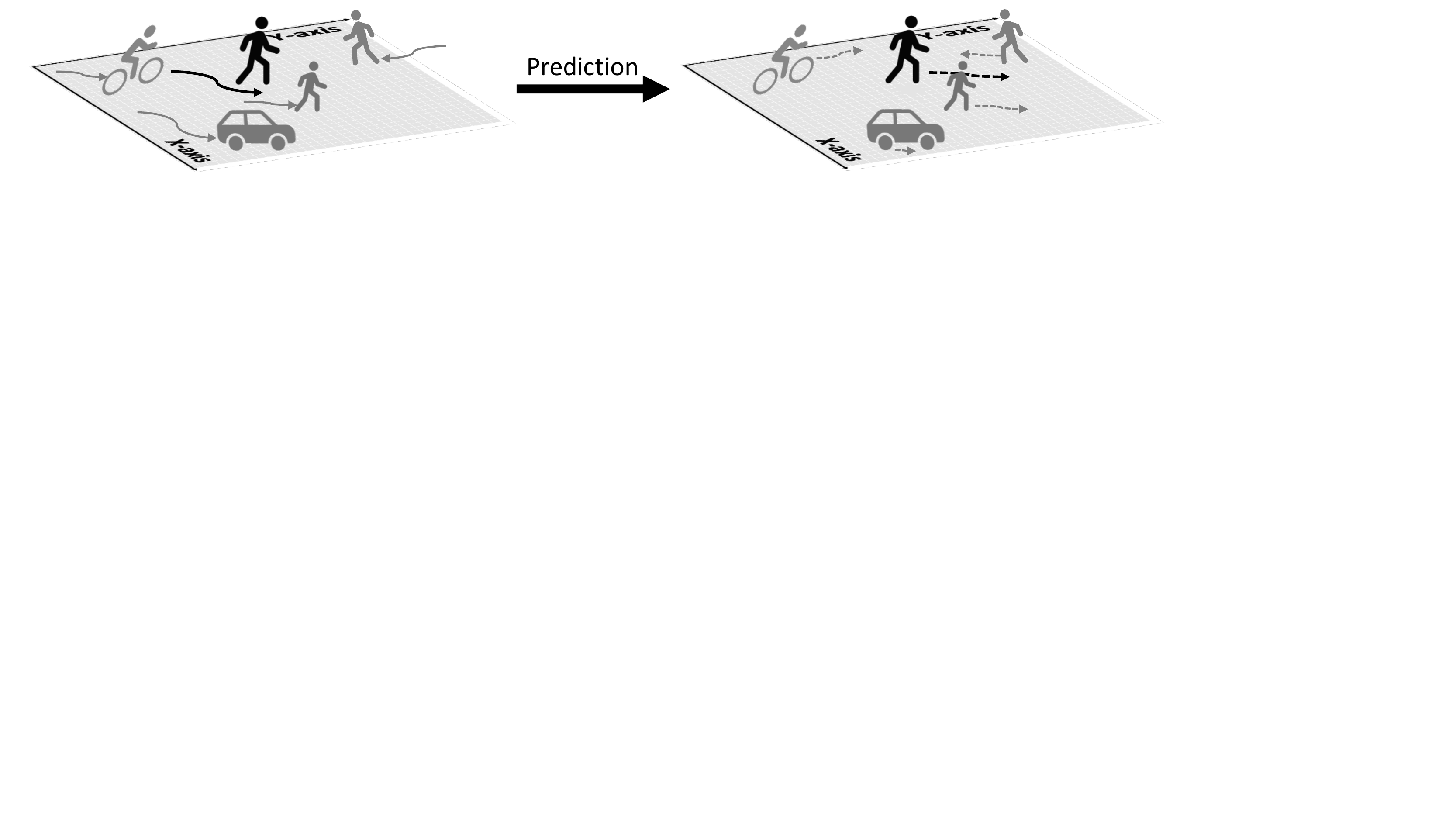}
    \caption{Predicting future positions of agents (\eg~target agent in black) at each  step within a predefined time interval by observing their past trajectories in mixed traffic situations.}
    \label{fig:sketch_map}
\end{figure}

Accurate trajectory prediction is a crucial task in different communities, such as intelligent transportation systems (ITS) for traffic management and autonomous driving~\cite{morris2008survey,cheng2018modeling,cheng2020mcenet}, 
photogrammetry mapping and extraction~\cite{schindler2010automatic,klinger2017probabilistic,cheng2018mixed,ma2019deep}, 
computer vision~\cite{alahi2016social,mohajerin2019multi} and mobile robot applications~\cite{mohanan2018survey}.
It enables an intelligent system to foresee the behaviors of road users and make a reasonable and safe decision for the next operation.
It is defined as the prediction of plausible (\eg~collision free and energy efficient) and socially-acceptable (\eg~considering social rules, norms, and relations between agents) positions in 2D/3D of non-erratic target agents (pedestrians, cyclists, vehicles and other types~\cite{rudenko2019human}) at each step within a predefined future time interval relying on observed partial trajectories over a certain period of discrete time steps~\cite{helbing1995social,alahi2016social}.
A  prediction process in mixed traffic is exemplified in Fig.~\ref{fig:sketch_map}. 

How to effectively predict accurate trajectories for heterogeneous agents still remains challenging due to: 1) the complex behavior and uncertain moving intention of each agent, 2) the presence and interactions between agents, and 3) multi-path choices: there is usually more than one socially-acceptable path that an agent could use in the future.

Boosted by Deep Learning (DL)~\cite{lecun2015deep} technologies and the availability of large-scale real-world datasets and benchmarks, recent methods utilizing Recurrent Neural Networks (RNNs) and/or Convolutional Neural Networks (CNNs) have made significant progress in modeling the interactions between agents and predicting their future trajectories ~\cite{alahi2016social,lee2017desire,vemula2018social,gupta2018social,xue2018ss,cheng2020mcenet}.
However, it is difficult for those methods to distinguish the effects of heterogeneous neighboring agents in different situations. For example, the target vehicle is affected more by the pedestrians in front of it tending to cross the road than by the following vehicles. 
Besides, minimizing the Euclidean distance between the ground truth and the prediction is commonly used as the objective function in some discriminative models~\cite{vemula2018social,xue2018ss}, which produce a deterministic outcome and is likely to predict the ``average" trajectories. 
In this regard, generative models \cite{goodfellow2014generative,kingma2014auto,kingma2014semi,sohn2015learning} are proposed for predicting multiple socially-acceptable trajectories~\cite{lee2017desire,gupta2018social}. 
In spite of the great progress, most of these methods are designed for homogeneous agents (\eg~only pedestrians). 
An important research question remains open: how to predict accurate trajectories in different scenes for all the various types of heterogeneous agents?

To address this problem, we propose a model named \emph{Attentive Maps Encoder Network} (AMENet). It inherits the ability of deep conditional generative models~\cite{sohn2015learning} using Gaussian latent variables for modeling complex future trajectories and learns the interactions between agents by attentive dynamic maps. 
The interaction module manipulates the information extracted from the neighboring agents' orientation, speed and position in relation to the target agent at each step and the attention mechanism~\cite{vaswani2017attention} enables the module to automatically focus on the salient features extracted over different steps. 
Fig.~\ref{fig:framework} gives an overview of the model.
Two encoders learn the representations of an agent's behavior into a latent space: the X-Encoder learns the information from the observed trajectories, while the Y-Encoder learns the information from the future trajectories of the ground truth and is removed in the inference phase.
The Decoder is trained to predict the future trajectories conditioned on the information learned by the X-Encoder and the representations sampled from the latent space.

The main contributions of this study are summarized as follows:
\begin{itemize}[nosep]
    \item[1] The generative framework AMENet encodes uncertainties of an agent's behavior into the latent space and predicts multi-path trajectories.
    \item[2] A novel module, \emph{attentive dynamic maps}, learns spatio-temporal interconnections between agents considering their orientation, speed and position.
    \item[3] It predicts accurate trajectories for heterogeneous agents in various unseen real-world environments, rather than focusing on homogeneous agents.
\end{itemize}

The efficacy of AMENet is validated using the recent benchmarks \emph{Trajnet}~\cite{sadeghiankosaraju2018trajnet} that contains 20 unseen scenes in various environments and InD~\cite{inDdataset} of four different intersections for trajectory prediction. Each module of AMENet is validated via a series of ablation studies. Its detailed implementation information is given in Appendix~C and the source code is available at \url{https://github.com/haohao11/AMENet}. 

\section{Related Work}
Trajectory prediction has been studied for decades and we discuss the most relevant works with respect to sequence prediction, interaction modeling and generative models for multi-path prediction.

\subsection{Sequence Modeling}
\label{sec:rel-seqmodeling}
Modeling trajectories as sequences is one of the most common approaches. The 2D/3D positions of an agent are predicted step by step.
The widely applied models include linear regression and Kalman filter~\cite{harvey1990forecasting}, Gaussian processes~\cite{tay2008modelling} and Markov decision processing~\cite{kitani2012activity}. 
These traditional methods largely rely on the quality of manually designed features and have limited performance in tackling large-scale data. 
Benefiting from the development of DL technologies \cite{lecun2015deep} in recent years, the RNNs and ~Long Short-Term Memories (LSTMs)~\cite{hochreiter1997long} are inherently designed for sequence prediction tasks and successfully applied for predicting pedestrian trajectories~\cite{alahi2016social,gupta2018social,sadeghian2018sophie,zhang2019sr} and other types of road users~\cite{mohajerin2019multi,chandra2019traphic,tang2019multiple}.
In this work, we use LSTMs to encode the temporal sequential information and decode the learned features to predict trajectories in sequence.

\subsection{Interaction Modeling}
\label{sec:rel-intermodeling}
The behavior of an agent can be crucially affected by others. Therefore, effectively modeling the interactions is important for accurate trajectory prediction.
The negotiation between road agents is simulated by Game Theory~\cite{johora2020agent} or Social Forces~\cite{helbing1995social}, \ie~the repulsive force for collision avoidance and the attractive force for social connections.
Such rule-based interaction modelings have been incorporated into DL models. Social LSTM~\cite{alahi2016social} proposes an occupancy grid to map the positions of close neighboring agents and uses a Social pooling layer to encode the interaction information for trajectory prediction. Many works design their specific ``occupancy'' grid~\cite{lee2017desire,xue2018ss,hasan2018mx,cheng2018modeling,cheng2018mixed,johora2020agent}.
Cheng~\etal~\cite{cheng2020mcenet} consider the interactions between individual and group agents with social connections and report better performance.
Meanwhile, different pooling mechanisms are proposed. The generative adversarial network (GAN)~\cite{goodfellow2014generative} based model Social GAN~\cite{gupta2018social} embeds relative positions between the target and all the other agents and then uses an element-wise pooling to extract the interaction between all the pairs of agents;
The SR-LSTM (States Refinement LSTM) model~\cite{zhang2019sr} proposes a states refinement module for aligning all the agents together and adaptively refines the state of each agent through a message passing framework. 
However, only the position information is leveraged in most of the above DL models.
The interaction dynamics are not fully captured both in spatial and temporal domains.

\subsection{Modeling with Attention}
\label{sec:rel-attention}
Attention mechanisms~\cite{bahdanau2014neural,xu2015show,vaswani2017attention} have been utilized to extract semantic information for predicting trajectories \cite{varshneya2017human,sadeghian2018sophie,al2018move,giuliari2020transformer}.
A soft attention mechanism~\cite{xu2015show} is incorporated in LSTMs to learn the spatio-temporal patterns from the position coordinates~\cite{varshneya2017human}. 
SoPhie~\cite{sadeghian2018sophie} applies two separate soft attention modules: the physical attention learns salient agent-to-scene features and the social attention models agent-to-agent interactions. 
In the MAP model (Move, Attend, and Predict)~\cite{al2018move}, an attentive network is implemented to learn the relationships between the location and time information. 
The most recent work Ind-TF~\cite{giuliari2020transformer} utilizes the Transformer network~\cite{vaswani2017attention} for modeling trajectory sequences. Transformer is a type of neural network structure for modeling sequences and widely applied in machine translation for sequence prediction.
In this work, we model the dynamic interactions among all road users by utilizing the self-attention mechanism~\cite{vaswani2017attention} along the time axis. 

\subsection{Generative Models}
\label{sec:rel-generative}
Nowadays, in the era of DL, GAN~\cite{goodfellow2014generative}, VAE~\cite{kingma2014auto} and the variants such as CVAE~\cite{kingma2014semi,sohn2015learning}, are the most popular generative models.
Gupta~\etal~\cite{gupta2018social} trained a generator to generate future trajectories from noise and a discriminator to judge whether the generated ones are fake or not. The performances of the two modules are enhanced mutually and the generator is able to generate trajectories that are as precise as the real ones. Amirian~\etal~\cite{amirian2019social} propose a GAN-based model for generating multiple plausible trajectories for pedestrians.
The CVAE model is used to predict multi-path trajectories conditioned on the observed trajectories~\cite{lee2017desire}, as well as scene context~\cite{cheng2020mcenet}.
Besides the generative models, Makansi \etal~\cite{makansi2019overcoming} treat the multi-path trajectory prediction problem as multi-model distributions estimation. Their method first predicts multi-model distributions with an evolving strategy by combining Winner-Takes-ALL loss~\cite{guzman2012multiple}, and then fits a distribution to the samples from the first stage for trajectory prediction.

In this work, we incorporate a CVAE module to learn a latent space for predicting multiple plausible future trajectories conditioned on the observed trajectories.
Our work essentially differs from the above models in the following ways. Interactions are modeled by the dynamic maps considering 1) not only position, but also orientation and speed and 2) are automatically extracted with the self-attention mechanism, and 3)  the interactions associated with the ground truth are also encoded into the latent space, which is different from a conventional CVAE model only ``auto-encoding" the ground truth trajectories~\cite{lee2017desire}.

\begin{figure}[t!]
    \centering
    \includegraphics[trim=0.5in 3.5in 0in 0.5in, width=1\textwidth]{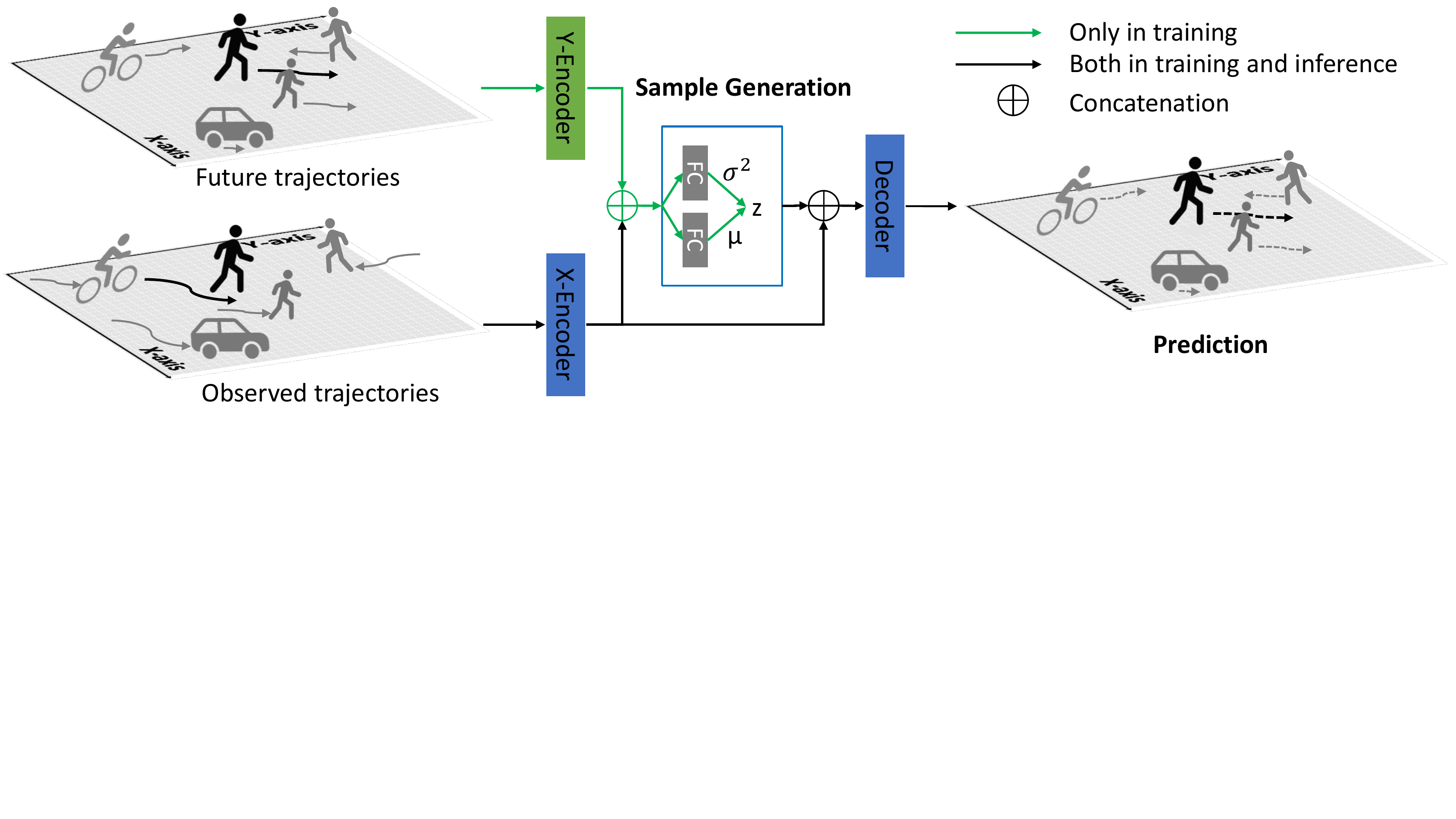}
    \caption{An overview of the proposed framework. It consists of four modules: the X-Encoder and Y-Encoder are used for encoding the observed and the future trajectories, respectively. They have an identical structure. The Sample Generator produces diverse samples conditioned on the input of the previous encoders. The Decoder is used to decode the features from the produced samples and predicts the future trajectories sequentially. FC stands for fully connected layer. The specific structure of the X-Encoder/Y-Encoder is given by Fig.~\ref{fig:encoder}.}
    \label{fig:framework}
\end{figure}

\section{Methodology}
\label{sec:methodology}

In this section, we introduce the proposed model AMENet (Fig.~\ref{fig:framework}) in detail in the following structure: a brief review on the \emph{CVAE} (Sec.~\ref{subsec:cvae}), the detailed structure of \emph{AMENet} (Sec.~\ref{subsec:AMENet}) and the \emph{Feature Encoding} (Sec.~\ref{subsubsec:featureencoding}) of the motion input and the attentive dynamic maps.

\subsection{Diverse Sample Generation with CVAE}
\label{subsec:cvae}

The CVAE model is an extension of the generative model VAE \cite{kingma2014auto} that introduces a condition to control the output \cite{kingma2014semi}. More details of the theory is provided in Appendix~A. The following describes the basics of the CVAE model.
Given a set of samples $(\boldsymbol{X, Y})=((\boldsymbol{X}_1, \boldsymbol{Y}_1 ),\cdots,(\boldsymbol{X}_N, \boldsymbol{Y}_N))$, it jointly learns a recognition model $q_\phi(\mathbf{z}|\boldsymbol{Y},\,\boldsymbol{X})$ of a variational approximation of the true posterior $p_\theta(\mathbf{z}|\boldsymbol{Y},\,\boldsymbol{X})$ and a generation model $p_\theta(\boldsymbol{Y}|\boldsymbol{X}, \,\boldsymbol{z})$ for predicting the output $\boldsymbol{Y}$ conditioned on the input $\boldsymbol{X}$. $\boldsymbol{z}$ are the stochastic latent variables, $\phi$ and $\theta$ are the respective recognition and generative parameters. The goal is to maximize the \textit{Conditional Log-Likelihood}:
\begin{equation}
\begin{split}
    \log{p_\theta(\boldsymbol{Y}|\boldsymbol{X})} &= \log\sum_{\boldsymbol{z}}  p_\theta(\boldsymbol{Y}, \boldsymbol{z}|\boldsymbol{X})\\
    &= \log{(\sum_{\boldsymbol{z}} q_\phi(\boldsymbol{z}|\boldsymbol{X}, \boldsymbol{Y})\frac{p_\theta(\boldsymbol{Y}|\boldsymbol{X}, \boldsymbol{z})p_\theta(\boldsymbol{z}|\boldsymbol{X})}{q_\phi(\boldsymbol{z}|\boldsymbol{X}, \boldsymbol{Y})})}.
\end{split}
\end{equation}
By means of Jensen's inequality, the evidence lower bound can be obtained:
\begin{equation}
\label{eq:CVAE}
    \log{p_\theta(\boldsymbol{Y}|\boldsymbol{X}}) \geq
    -D_{KL}(q_\phi(\mathbf{z}|\boldsymbol{X}, \,\boldsymbol{Y})||p_\theta(\mathbf{z}))
    + \mathbb{E}_{q_\phi(\mathbf{z}|\boldsymbol{X}, \,\boldsymbol{Y})}
    [\log p_\theta(\boldsymbol{Y}|\boldsymbol{X}, \,\mathbf{z})].
\end{equation}
Here both the approximate posterior $q_\phi(\mathbf{z}|\boldsymbol{X}, \,\boldsymbol{Y})$ and the prior $p_\theta(\mathbf{z})$ are assumed to be Gaussian distributions for an analytical solution \cite{kingma2014auto}. During training, the Kullback-Leibler divergence $D_{KL}(\cdot)$ pushes the approximate posterior to the prior distribution $p_\theta(\mathbf{z})$. The generation error $\mathbb{E}_{q_\phi(\mathbf{z}|\boldsymbol{X}, \,\boldsymbol{Y})}(\cdot)$ measures the distance between the generated output and the ground truth. During inference, for a given observation $\boldsymbol{X}_i$, one latent variable $z$ is drawn from the prior distribution $p_\theta(\mathbf{z})$, and one of the possible outputs $\hat{\boldsymbol{Y}}_i$ is generated from the distribution $p_\theta(\boldsymbol{Y}_i|\boldsymbol{X}_i,\,z)$. The latent variables $\mathbf{z}$ allow for the one-to-many mapping from the condition to the output via multiple sampling. 

\subsection{Attentive Encoder Network for Trajectory Prediction}
\label{subsec:AMENet}
In tasks like trajectory prediction, we are interested in modeling a conditional distribution $p_\theta(\boldsymbol{Y}_n|\boldsymbol{X})$, where $\boldsymbol{X}$ is the past trajectory information and $\boldsymbol{Y}_n$ is one of its possible future trajectories.
In order to realize this goal that generates controllable samples of future trajectories based on past trajectories, a CVAE module is adopted inside our framework.

The multi-path trajectory prediction problem with the consideration of motion and interaction information is defined as follows: agent $i$, receives as input its observed trajectories $\boldsymbol{X}_i=\{X_i^1,\cdots,X_i^T\}$ for predicting its $n$-th plausible future trajectory $\hat{\boldsymbol{Y}}_{i,n}=\{\hat{Y}_{i,n}^1,\cdots,\hat{Y}_{i,\,n}^{T'}\}$. $T$ and $T'$ denote the total number of steps of the past and future trajectories, respectively. The trajectory position of $i$ is characterized by the coordinates as $X_i^t=({x_i}^t, {y_i}^t)$ at step $t$ or as $\hat{Y}_{i,n}^{t'}=(\hat{x}_{i,n}^{t'}, \hat{y}_{i,n}^{t'})$ at step $t'$. 3D coordinates are also possible, but in this work only 2D coordinates are considered. The objective is to predict its multiple plausible future trajectories $( \hat{\boldsymbol{Y}}_{i,1},\cdots,\hat{\boldsymbol{Y}}_{i,N})$ that are as accurate as possible to the ground truth $\boldsymbol{Y}_i$. This task is mathematically defined as $\hat{\boldsymbol{Y}}_{i,\,n} = f(\boldsymbol{X}_i, \text{AMap}_i)$ and \rw{$n \leq N$}. Here, $N$ denotes the total number of the predicted trajectories and $\text{AMap}_i$ denotes the attentive dynamic maps centralized on the target agent for mapping the interactions with its neighboring agents over the steps. More details of the attentive dynamic maps will be given in Sec.~\ref{subsubsec:dynamic}. 

We extend the CVAE model as follows to solve this problem:
\begin{equation}
 \begin{split}
            f(\boldsymbol{X}_i, \text{AMap}_i) &= \log{p_\theta(\boldsymbol{Y}_i|\boldsymbol{X}_i}, \text{AMap}_i),\\
                                      &\geq -D_{KL}(q_\phi(\mathbf{z}|\boldsymbol{X}_i \,\boldsymbol{Y}_i,\,\text{AMap}_i)||p_\theta(\mathbf{z}))\\
    &+ \mathbb{E}_{q_\phi(\mathbf{z}|\boldsymbol{X}_i,\,\boldsymbol{Y}_i,\,\text{AMap}_i)}
    [\log p_\theta(\boldsymbol{Y}_i|\boldsymbol{X}_i, \text{AMap}_i,\,\mathbf{z})].
 \end{split}  
\end{equation}
Note that for simplicity, the notation of steps $T$ and $T'$ is omitted.  $q_\phi(\cdot)$ accesses the interactions captured by $(\text{AMap}_i)_{t=1}^{T}$ and $(\text{AMap}_i)_{t'=1}^{T'}$, respectively, from both the observation and the future time, while $p_\theta(\cdot)$ only accesses the interactions captured by $(\text{AMap}_i)_{t=1}^{T}$ from the observation time.

In the training phase, $q_\phi(\cdot)$ and $p_\theta(\cdot)$ are jointly learned. The recognition model is trained via the X-Encoder and Y-Encoder. 
The encoded outputs from both encoders are concatenated and then forwarded to two side-by-side fully connected (FC) layers to produce the mean and the standard deviation of the latent variables $\mathbf{z}$. The generation model is trained via the Decoder. It takes the output of the X-Encoder as the condition and the latent variables to generate the future trajectory. We employ an LSTM network in the Decoder for predicting the future trajectory sequentially. 
The Mean Squared Error (MSE) between the predicted trajectories and the ground-truth ones is used as the reconstruction loss. During inference, the Y-Encoder is removed and the X-Encoder works in the same way as in the training phase. The Decoder generates a prediction conditioned on the output of the X-Encoder and the  sampled latent variable $z$. This step is repeated $N$ times to predict multiple trajectories.

\begin{figure}[t!]
    \centering
    \includegraphics[trim=0.5in 2.2in 3.6in 0.5in, width=0.75\textwidth]{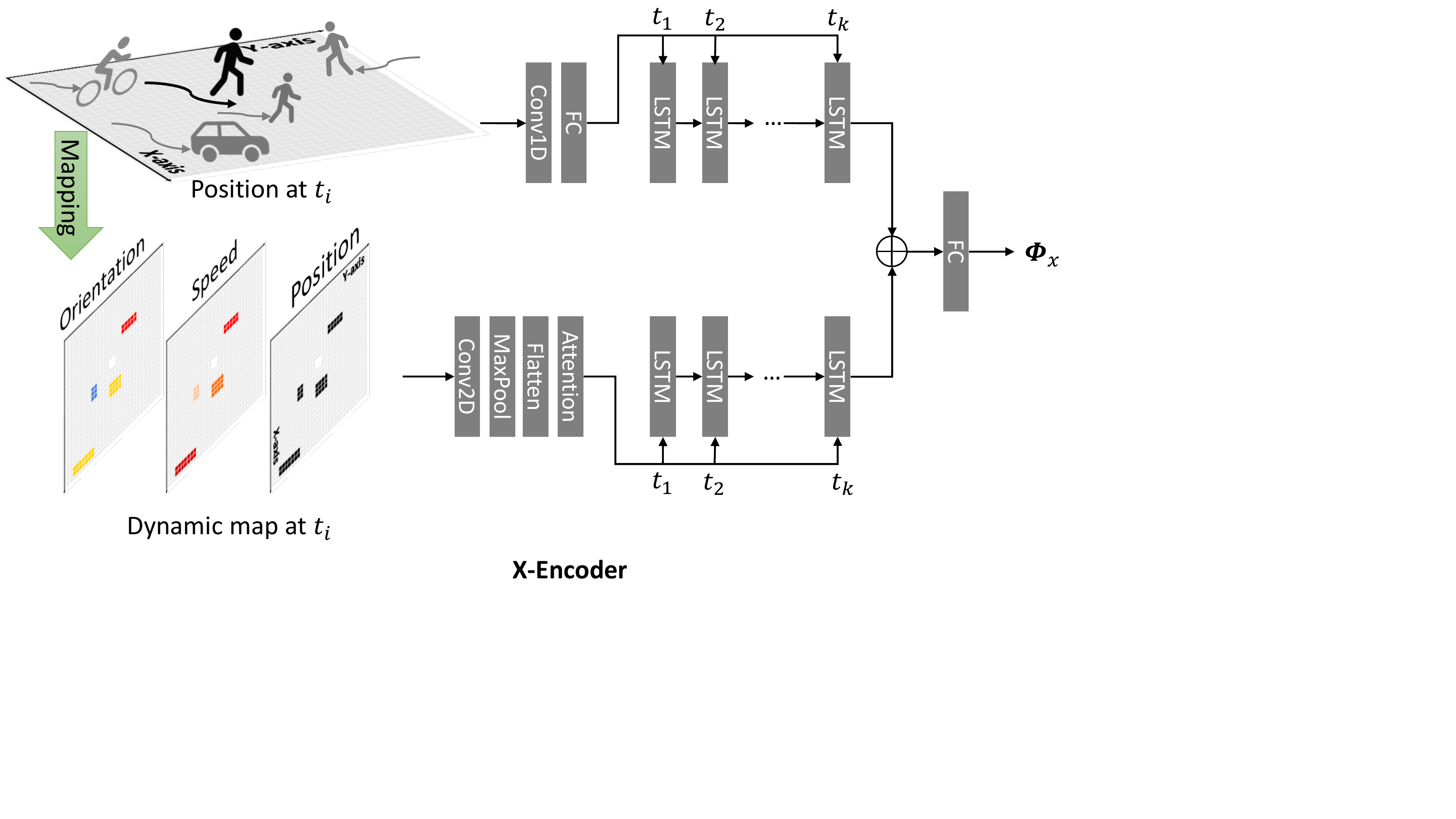}
    \caption{The structure of the X-Encoder. The upper branch extracts motion information of target agents and the lower one learns the interaction information between neighboring agents from the dynamic maps over time attentively. The motion information and the interaction information are encoded by their respective LSTMs sequentially. The last outputs of the two LSTMs are concatenated and forwarded to a fully connected (FC) layer to get the final output of the X-Encoder. The Y-Encoder has the same structure as the X-Encoder.}
    \label{fig:encoder}
\end{figure}

Fig.~\ref{fig:encoder} shows the detailed structure of the X-Encoder/Y-Encoder, which are designed for learning the information from the motion input and the attentive dynamic maps. 
The X-Encoder is used to encode the past information. It has two branches in parallel to process the motion information (upper branch) and dynamic maps information for interaction (lower branch). 
The upper branch takes the offsets $({\Delta{x}_i}^t, {\Delta{y}_i}^t)_{t=1}^{T}$ for each target agent over the observed steps. The motion information firstly is passed to a 1D convolutional layer (Conv1D) with a one-step stride along the time axis to learn motion features one step after another. Then the output is sequentially passed to a FC layer and an LSTM module for encoding the temporal features into a hidden state, which contains all the motion information of the target agent.
The lower branch takes the dynamic maps $({\text{Map}}_i^t)_{t=1}^T$ as input.
The interaction information at each step is passed through a 2D convolutional layer (Conv2D) with the ReLU activation and a Max Pooling layer (MaxPool) for learning the spatial features among all the agents. The output of MaxPool at each step is flattened and concatenated along the time axis to form a timely distributed feature vector. Then, the feature vector is fed forward to the attention layer for learning the interaction information. 
The output of the attention layer is passed to an LSTM  used to encode the dynamic interconnection in the sequence.
Both the hidden states (the last output) from the motion LSTM and the interaction LSTM are concatenated and passed to a FC layer for feature fusion, as the complete output of the X-Encoder.

The Y-Encoder has the same structure as the X-Encoder, which is used to encode both the target agent's motion and interaction information from the ground truth during the training time.  The dynamic maps are also leveraged in the Y-Encoder, although, they are not reconstructed from the Decoder (only the future trajectories are reconstructed). This extended structure distinguishes our model from the conventional CVAE structure~\cite{kingma2014auto,kingma2014semi,sohn2015learning} and the work from~\cite{lee2017desire}, in which only the ground truth trajectories are inserted for training the recognition model (see Sec.~\ref{subsec:cvae}).

For tasks of single-path prediction, such as the Trajnet challenge or path planning, a ranking strategy is proposed to select the \textit{most-likely} predicted trajectory out of the multiple predictions. 
We apply a bivariate Gaussian distribution to rank the predicted trajectories $(\hat{\boldsymbol{Y}}_{i,1},\cdots,\hat{\boldsymbol{Y}}_{i,N})$ for each agent. At step $t'$, all the predicted positions for the agent $i$ are stored in the vector $|{\hat{\mathsf{X}}_{i}}, {\hat{\mathsf{Y}}_{i}}|^{t'}$. We follow the work ~\cite{graves2013generating} to fit the positions into a bivariate Gaussian distribution: 
\begin{equation}
\label{eq:ped}
    f(\hat{x}_i, \hat{y}_i)^{t'} = \frac{1}{2\pi\mu_{\hat{\mathsf{X}}_{i}}\mu_{\hat{\mathsf{Y}}_{i}}\sqrt{1-\rho^2}}\exp{\frac{-Z}{2(1-\rho^2)}},
\end{equation}
where
\begin{equation}
    Z = \frac{(\hat{x}_i-\mu_{\hat{\mathsf{X}}_{i}})^2}{{\sigma_{\hat{\mathsf{X}}_{i}}}^2} + \frac{(\hat{y}_i-\mu_{\hat{\mathsf{Y}}_{i}})^2}{{\sigma_{\hat{\mathsf{Y}}_{i}}}^2} -  \frac{2\rho(\hat{x}_i-\mu_{\hat{\mathsf{X}}_{i}})(\hat{y}_i-\mu_{\hat{\mathsf{Y}}_{i}})}{\sigma_{\hat{\mathsf{X}}_{i}}\sigma_{\hat{\mathsf{Y}}_{i}}}.
\end{equation}
$\mu$ denotes the mean and $\sigma$ the standard deviation. $\rho$ is the correlation between $\hat{\mathsf{X}}_{i}$ and $\hat{\mathsf{Y}}_{i}$.
A predicted trajectory is scored as the sum of the relative likelihood of all of its steps:
\begin{equation}
    S(\boldsymbol{\hat{Y}}_{i,n}) = \sum_{t'=1}^{T'}f(\hat{x}_i, \hat{y}_i)^{t'}.
\end{equation}
All the predicted trajectories are ranked according to this score. The one with the highest score is selected for the single-path prediction.

\subsection{Feature Encoding}
\label{subsubsec:featureencoding}
In this subsection we discuss how to encode the motion and interaction information in detail.

\subsubsection{Motion Input} 
\label{subsubsec:input}
The motion information for each agent is captured by the position coordinates at each step.
Specifically, we use the offset of the trajectory positions between two consecutive steps $({\Delta{x}}^t, {\Delta{y}}^t) = (x^{t+1} - x^{t}, ~ y^{t+1} - y^{t}) $ as the motion information, which has been widely applied in this domain \cite{gupta2018social,becker2018evaluation,zhang2019sr,cheng2020mcenet}.
Compared to coordinates, the offset is independent from the given space and less sensitive with respect to overfitting a model to particular space or travel directions.   
It is interpreted as speed over steps that are defined with a constant duration. 
The coordinates at each position are calculated back by cumulatively summing the sequence offsets from the given original position. 
The class information of agent's type is useful for analyzing its motion \cite{cheng2020mcenet}. However, the Trajnet benchmark does not provide this information for the trajectories and we do not use it here.
As augmentation technique we randomly rotate the trajectories to prevent the model from only learning certain directions. In order to maintain the relative positions and angles between agents, the trajectories of all the agents coexisting in a given period are rotated by the same angle.

\subsubsection{Attentive Dynamic Maps}
\label{subsubsec:dynamic}
We propose a novel and straightforward method: attentive dynamic maps to learn agent-to-agent interaction information. The mapping method is inspired by the recent works of parsing the interactions between agents based on an occupancy grid \cite{alahi2016social,lee2017desire,xue2018ss,hasan2018mx,cheng2018modeling,cheng2018mixed,johora2020agent}, which uses a binary tensor to map the relative positions of the neighboring agents of the target agent \cite{alahi2016social}. The so-called dynamic maps extend this method, in which the interactions at each step are modeled via three layers dedicated for orientation, speed and position information. Each map changes from one step to the next and therefore the spatio-temporal interaction information between agents is interpreted dynamically over time.

The map is defined as a rectangular area around the target agent, and is divided into grid cells and centralized on the agent's current location, see Fig.~\ref{fig:encoder}. $W$ and $H$ denote the width and height. 
First, referred to the target agent $i$, the neighboring agents $\mathsf{N}(i)$ are mapped into the closest grid $\text{cells}_{w \times h}^t$ according to their relative position and they are also mapped onto the cells reached by their anticipated relative offset (speed) in the $x$ and $y$ directions. 
\begin{equation}
\label{eq:cell}
\begin{split}
  &\text{cells}_{w}^t = x_j^t-x_i^t + (\Delta x_j^t - \Delta{x}_i^t), \\
  &\text{cells}_{h}^t  = y_j^t-y_i^t + (\Delta y_j^t - \Delta{y}_i^t), \\
  &\text{where~} w \leq W ,~ h \leq H,~ j \in \mathsf{N}(i)~\text{and}~ j\neq i.
\end{split}
\end{equation}
Second, the orientation, speed and position information is stored in the mapped cells in the respective layer for each neighboring agent. The \emph{orientation} layer $O$ stores the heading direction. For the neighboring agent $j$, its orientation from the current to the next position is the angle $\vartheta_j$ in the Euclidean plane and calculated in the given radians by $\vartheta_j = \text{arctan2} (\Delta y_j^t,\, \Delta x_j^t)$. Its value is shifted into the interval $[0^{\circ},\,360^{\circ})$. 
Similarly, the \emph{speed} layer $S$ stores the travel speed and the \emph{position} layer $P$ stores the positions using a binary flag in the cells mapped above. Last, layer-wise, a Min-Max normalization scheme is applied for normalization.

The map covers a large vicinity area. Empirically we found $32\times32\,\text{m}^2$ a proper setting considering both the coverage and the computational cost. There is a trade-off between a high and a low resolution map. A cell is filled by a maximum of one agent if its size is small. But the high resolution will lead to a very sparse map (most of the cells have no value) and the surrounding areas of the neighboring agent will be treated as having no impact on the target agent. On the other hand, there may exist an overlap of multiple agents in one cell with a very different travel speed or orientation if the cell size is too big. In this study we resolve this problem by setting the cell size 
as $1\times1\,\text{m}^2$. Based on the distribution of the experimental data, there are only a few cells with overlapped agents, which is also supported by the preservation of personal space~\cite{gerin2005negotiation}. However, the information of agents' size is not given by the experimental data, the approximation of the cell size may not be valid for large agents. In future work, the size of the agents will be considered and an extended margin will be applied to avoid the problem of agents falling out of the grid cell bounds.

Interactions among different agents are dynamic in various situations from one step to another in a sequence. Some steps may impact the agents' behaviors more than other steps.
To explore such varying information, we employ a self-attention mechanism~\cite{vaswani2017attention} to learn the interaction information from the dynamic maps over time and call them \emph{attentive dynamic maps}. The self-attention module takes as input the dynamic maps and attentively learns the interconnections over the steps. The detailed information of this module is given in Appendix~B.

\section{Experiments}
\label{sec:experiments}
In this section, the evaluation metrics, benchmarks, experimental settings and the recent state-of-the-art methods for comparison are introduced for evaluating the proposed model. The ablation studies that partially remove the modules in the proposed method are conducted to justify each module's contribution. Finally, the experimental results are analyzed and discussed in detail.

\subsection{Evaluation Metrics}
The mean average displacement error (ADE) and final displacement error (FDE) are the most commonly applied metrics to measure the performance of trajectory prediction~\cite{alahi2016social,gupta2018social,sadeghian2018sophie}. ADE is the aligned Euclidean distance from $Y$ (ground truth) to its prediction $\hat{Y}$ averaged over all steps. FDE is the Euclidean distance of the last position from $Y$ to the corresponding $\hat{Y}$. It measures a model's ability for predicting the destination and is more challenging as errors accumulate with time. We report the mean values for all the trajectories.

We evaluate both the most-likely prediction and the best prediction $@top10$ for the multi-path trajectory prediction.
The most-likely prediction is selected by the trajectories ranking as described in Sec~\ref{subsec:AMENet}.
$@top10$ prediction is the best one out of ten predicted trajectories that has the smallest ADE and FDE compared with the ground truth. When the ground truth is not available (for the online test), only the most-likely prediction is selected. Then it becomes to the single trajectory prediction problem, as most of the previous works did~\cite{helbing1995social,alahi2016social,zhang2019sr,becker2018evaluation,hasan2018mx,giuliari2020transformer}.

\subsection{Trajnet Benchmark Challenge}
\label{subsec:benchmark}
We first verify the performance of the proposed method on Trajnet~\cite{sadeghiankosaraju2018trajnet}. It is the most popular large-scale trajectory-based activity benchmark in this domain and provides a uniform evaluation system for fair comparison among different submitted methods. A wide range of datasets (\eg~ETH~\cite{pellegrini2009you}, UCY~\cite{lerner2007crowds} and Stanford Drone Dataset~\cite{robicquet2016learning}) for heterogeneous agents (pedestrians, bikers, skateboarders, cars, buses, and golf cars) that navigate in real-world outdoor mixed traffic environments are included.
The data was collected from 38 scenes with ground truth for training and another 20 scenes without ground truth for testing (\ie~open challenge competition). Each scene presents various traffic densities in different space layouts, which makes the prediction task challenging and requires a model to generalize, in order to adapt to the various complex scenes. Trajectories are provided as the $xy$ coordinates in meters (or pixels) projected on a Cartesian space, with 8 steps for observation and the following 12 steps for prediction. 
The duration between two successive steps is 0.4 seconds. 
We follow all the previous works~\cite{helbing1995social,alahi2016social,zhang2019sr,becker2018evaluation,hasan2018mx,gupta2018social,giuliari2020transformer} that use the coordinates in meters. 

\begin{figure}[bpht!]
    \centering
    \includegraphics[width=\textwidth]{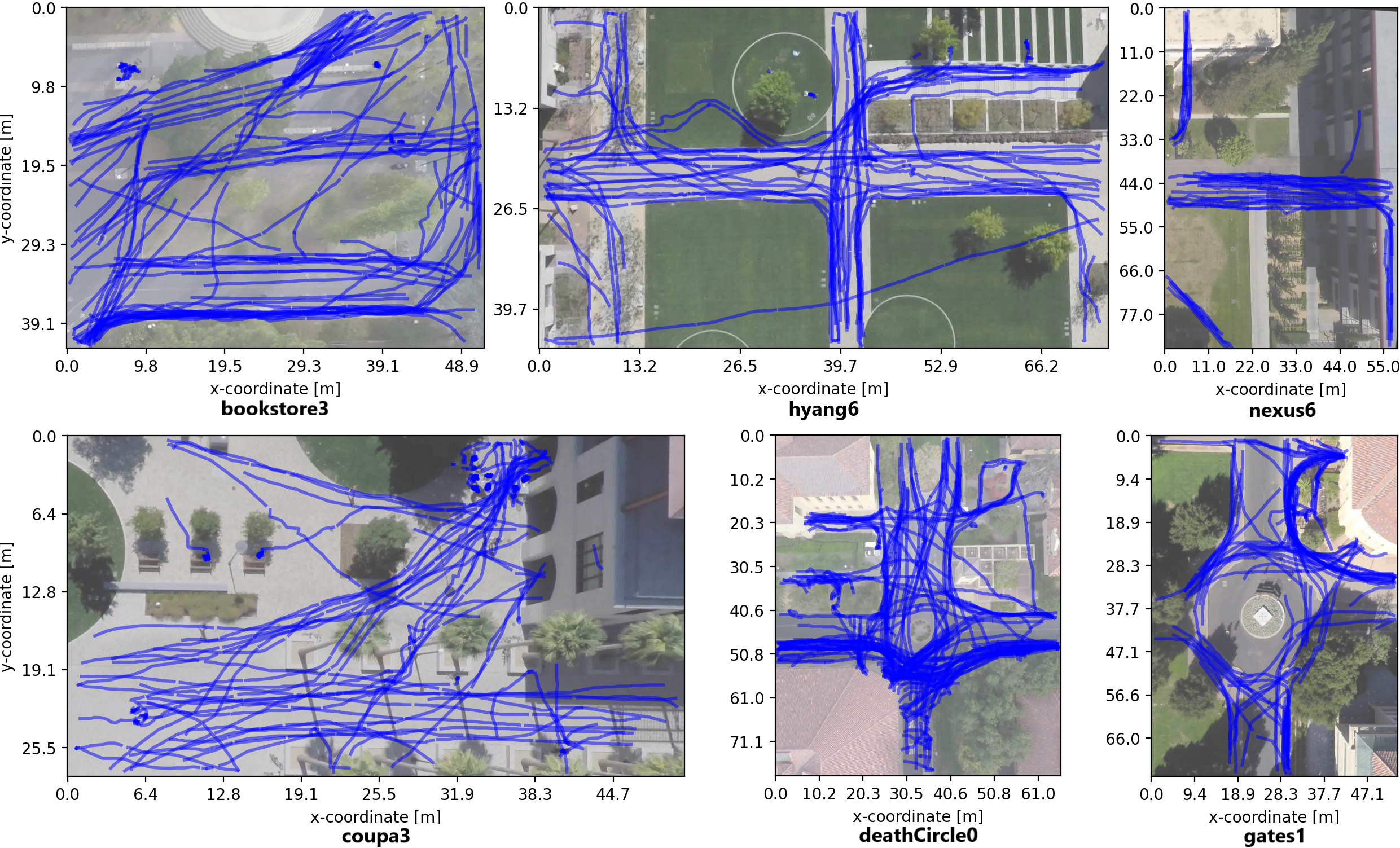}%
     \label{trajectories_nexus_0}
    \caption{Visualization of each scene of the offline test set.}
    \label{fig:trajectories}
\end{figure}

In order to train and evaluate the proposed method, as well as the ablation studies,
6 of the total 38 scenes in the training set are selected as the offline test set. Namely, they are \textit{bookstore3}, \textit{coupa3}, \textit{deathCircle0}, \textit{gates1}, \textit{hyang6}, and \textit{nexus0}. 
The selection of the scenes is based on the space layout, data density and percentage of non-linear trajectories, see Table~\ref{tb:multipath}. Fig.~\ref{fig:trajectories} visualizes the trajectories in each scene.
The trained model that has the best performance on the offline test set is selected as our final model and used for the online testing.

\subsection{Quantitative Results and Comparison}
\label{subsec:results}
We compare the performance of our model with the most influential previous works and the recent state-of-the-art works published on the Trajnet challenge.

\begin{itemize}[leftmargin=*,nosep]
\item\emph{Social Force}~\cite{helbing1995social} is a rule-based model with the repulsive force for collision avoidance and the attractive force for social connections; 
\item\emph{Social LSTM}~\cite{alahi2016social} proposes Social pooling with a rectangular occupancy grid for close neighboring agents, which is widely adopted in this domain~\cite{lee2017desire,xue2018ss,hasan2018mx,cheng2018modeling,cheng2018mixed,johora2020agent}; 
\item\emph{SR-LSTM}~\cite{zhang2019sr} uses a states refinement module for extracting social effects between the target agent and its neighboring agents; 
\item\emph{RED}~\cite{becker2018evaluation}  uses RNN-based Encoder with Multilayer Perceptron (MLP) for trajectory prediction; 
\item\emph{MX-LSTM}~\cite{hasan2018mx} exploits the head pose information of agents to help analyze its moving intention;
\item\emph{Social GAN}~\cite{gupta2018social} proposes to utilize GAN for multi-path trajectory prediction, which is the one of the closest works to our work; the other one is DESIRE~\cite{lee2017desire}. But neither the online test nor code was reported. Hence, we do not compare with DESIRE;
\item\emph{Ind-TF}~\cite{giuliari2020transformer} proposes a novel idea that utilizes the Transformer network~\cite{vaswani2017attention} for sequence prediction. No social interactions between agents are considered in this work.
\end{itemize}

The performances of single trajectory prediction from different methods on the Trajnet challenge are given in Table~\ref{tb:results}.
The results were originally reported on the leader board\footnote{\url{http://trajnet.stanford.edu/result.php?cid=1}} up to the date of 14 June 2020. AMENet outperformed the other models and won the first place measured by the aforementioned metrics. Compared with the most recent model Ind-TF~\cite{giuliari2020transformer}, AMENet achieved comparative performance in ADE and slightly better in FDE (from 1.197 to 1.183 meters). 
The superior performance given by AMENet here also validates the efficacy of the ranking method to select the most-likely prediction from the multiple predicted trajectories, as introduced in Sec.~\ref{subsec:AMENet}.

\begin{table}[t!]
\centering
\caption{Comparison between our method and the state-of-the-art models. Smaller values indicate a better performance and best values are highlighted in boldface.}
\begin{tabular}{llll}
\hline
Model                & Avg. [m]$\downarrow$ & FDE [m]$\downarrow$   & ADE [m]$\downarrow$ \\
\hline
Social LSTM~\cite{alahi2016social}         & 1.3865          & 3.098 & 0.675  \\
Social GAN~\cite{gupta2018social}           & 1.334           & 2.107 & 0.561  \\
MX-LSTM~\cite{hasan2018mx}              & 0.8865          & 1.374 & 0.399  \\
Social Force~\cite{helbing1995social}         & 0.8185          & 1.266 & 0.371  \\
SR-LSTM~\cite{zhang2019sr}              & 0.8155          & 1.261 & 0.370  \\
RED~\cite{becker2018evaluation}                  & 0.78            & 1.201 & 0.359  \\
Ind-TF~\cite{giuliari2020transformer}               & 0.7765          & 1.197 & \textbf{0.356}  \\
This work (AMENet)$^{*}$                 & \textbf{0.7695}          & \textbf{1.183} & \textbf{0.356}   \\ 
\hline
\end{tabular}
\begin{tabular}{@{}c@{}}
\multicolumn{1}{p{\textwidth}}{$^{*}$named as \textit{ikg\_tnt} on the leader board}
\end{tabular}
\label{tb:results}
\end{table}

\subsection{Results for Multi-Path Prediction}
\label{subsec:multipath-selection}

The performance for multi-path prediction is investigated using the offline test set. 
Table~\ref{tb:multipath} shows the quantitative results. Compared to the most-likely prediction, as expected the $@\text{top}10$ prediction yields similar but slightly better performance. It indicates that: (1) the generated multiple trajectories increase the chance to narrow down the errors; (2) the ranking method is effective for ordering the multiple predictions and proposing a good one, which is especially important when the prior knowledge of the ground truth is not available.

\begin{table}[t!]
\centering
\caption{Evaluation measured by ADE/FDE for multi-path trajectory prediction using AMENet on the Trajnet offline test set.} 
\begin{tabular}{llllll}
\hline
Dataset      & Layout       & \#Trajs & \begin{tabular}[c]{@{}l@{}}Non-linear \\ traj rate\end{tabular}   
                                           & @top10          & Most-likely       \\ \hline
bookstore3   & parking      & 429 & 0.71   & 0.477/0.961   & 0.486/0.979     \\
coupa3       & corridor     & 639 & 0.31   & 0.221/0.432   & 0.226/0.442     \\
deathCircle0 & roundabout   & 648 & 0.89   & 0.650/1.280   & 0.659/1.297     \\
gates1       & roundabout   & 268 & 0.87   & 0.784/1.663   & 0.797/1.692     \\
hyang6       & intersection & 327 & 0.79   & 0.534/1.076   & 0.542/1.094     \\
nexus6       & corridor     & 131 & 0.88   & 0.542/1.073   & 0.559/1.109     \\
Avg.         & -            & 407 & 0.74   & 0.535/1.081   & 0.545/1.102     \\ \hline
\end{tabular}
\label{tb:multipath}
\end{table}

\begin{figure}[t!]
    \centering
    \includegraphics[trim=0cm 0cm 0cm 0cm, width=0.475\textwidth]{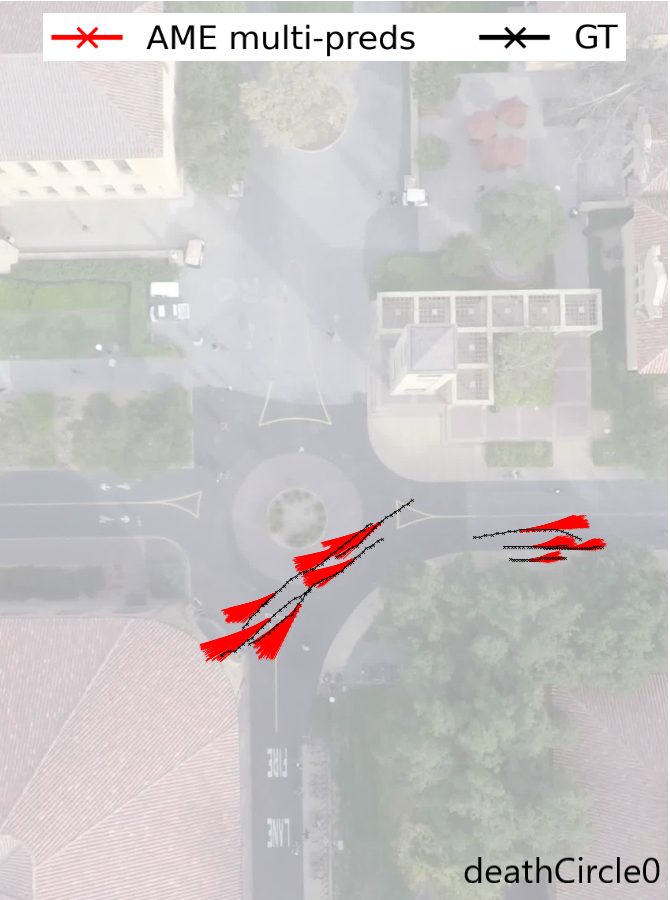}
    \includegraphics[trim=0cm 0cm 0cm 0cm, width=0.475\textwidth]{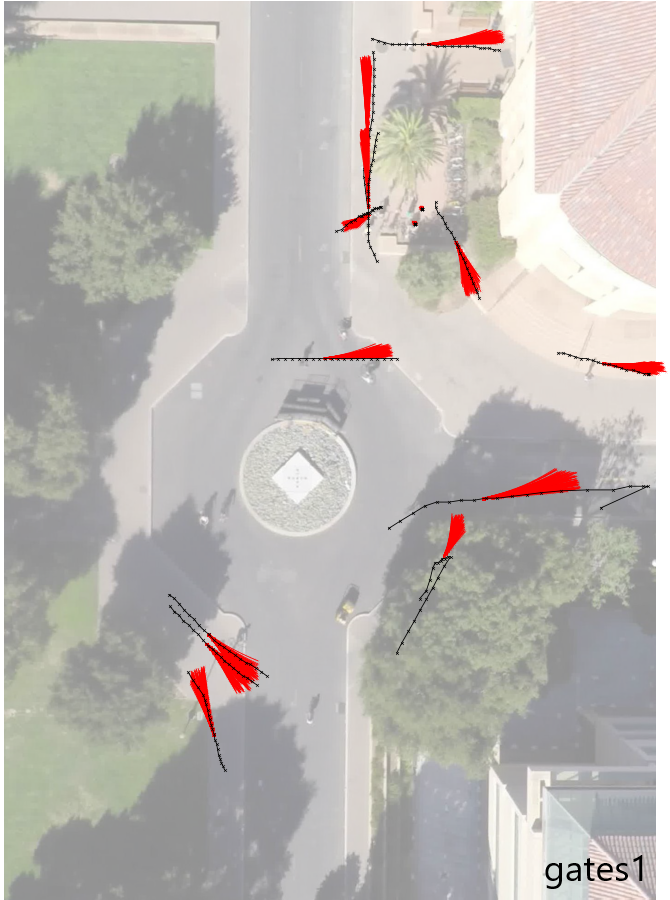}
    \caption{Multi-path predictions from AMENet.}
    \label{fig:multi-path}
\end{figure}

Fig.~\ref{fig:multi-path} showcases some qualitative examples of the multi-path trajectory prediction by AMENet. As shown in the roundabout deathCircle0 and gates1, each moving agent has more than one possibility (different speeds and orientations) to choose its future path. The predicted trajectories diverge more widely in further steps as the  uncertainty about an agent's intention increases with time. Predicting multiple plausible trajectories indicates a larger intended area and raises the chance to cover the path an agent might choose in the future. Also, the ``fan" of possible trajectories can be interpreted as reflecting the uncertainty of the prediction. Conversely, a single prediction provides limited information for inference and is likely to lead to a false conclusion if the prediction is not correct/precise in the early steps. On the other hand, agents that stand still were correctly predicted by AMENet with high certainty, as shown by two agents in gates1 in the upper right area. As designed by the model, only interactions between agents lead to adaptions in the predicted path and deviation from linear paths; the scene context, \eg~road geometry, is not modeled and thus does not affect prediction. 

\subsection{Ablation study}
\label{sec:ablativemodels}

In order to analyze the impact of each module in the proposed framework, \ie~dynamic maps, self-attention, and the extended structure of the CVAE, several ablative models were investigated.
\begin{itemize}[leftmargin=*,nosep]
    \item ENet: (E)ncoder (Net)work, which is only conditioned on the motion information. The interaction information is not leveraged. This model is treated as the baseline model.
    \item OENet: (O)ccupancy$+$ENet, where interactions are modeled by the occupancy grid  \cite{alahi2016social,lee2017desire,xue2018ss,hasan2018mx,cheng2018modeling,cheng2018mixed,johora2020agent} in both the X-Encoder and the Y-Encoder.
    \item AOENet: (A)ttention$+$OENet, where the self-attention mechanism is added.
    \item MENet: (M)aps$+$ENet, where interactions are modeled by the proposed dynamic maps in both the X-Encoder and the Y-Encoder.
    \item ACVAE: (A)ttention+CVAE, where the dynamic maps are only added in the X-Encoder. It is equivalent to a CVAE model~\cite{kingma2014auto,kingma2014semi,sohn2015learning} with the self-attention mechanism.
    \item AMENet: (A)ttention$+$MENet, where the self-attention mechanism is added. It is the full model of the proposed framework.
\end{itemize}
Table~\ref{tb:resultsablativemodels} shows the quantitative results for the ablation studies.
Errors are measured by ADE/FDE on the most-likely prediction. The comparison between OENet and the baseline model ENet shows that extracting the interaction information from the occupancy grid did not contribute to a better performance. Even though the self-attention mechanism was added to the occupancy grid (denoted by AOENet), the slightly enhanced performance still fell behind the baseline model. The comparison indicates that interactions were not effectively learned from the occupancy map with or without the self-attention mechanism across the datasets. The comparison between MENet and ENet shows a similar pattern. The performance was slightly less inferior using the dynamic maps than the occupancy grid (MENet vs. OENet) in comparison to the baseline model. However, profound improvements can be seen after employing the self-attention mechanism. First, the comparison between ACVAE and ENet shows that even without the extended structure in the Y-Encoder, the dynamic maps with the self-attention mechanism in the X-Encoder were very beneficial for modeling interactions. On average, the performance was improved by \SI{4.0}{\percent} and \SI{4.5}{\percent} as measured by ADE and FDE, respectively. Second, the comparison between the proposed model AMENet and ENet shows that after extending the dynamic maps to the Y-Encoder, the errors, especially the absolute values of FDE, further decreased across all the datasets; ADE was reduced by \SI{9.5}{\percent} and FDE was reduced by \SI{10.0}{\percent}. This improvement was also confirmed by the benchmark challenge (see Table~\ref{tb:results}).

\begin{figure}[t!]
    \centering
    \begin{subfigure}{0.5\textwidth}
    \includegraphics[trim=0cm 0cm 0cm 0cm, width=1\textwidth]{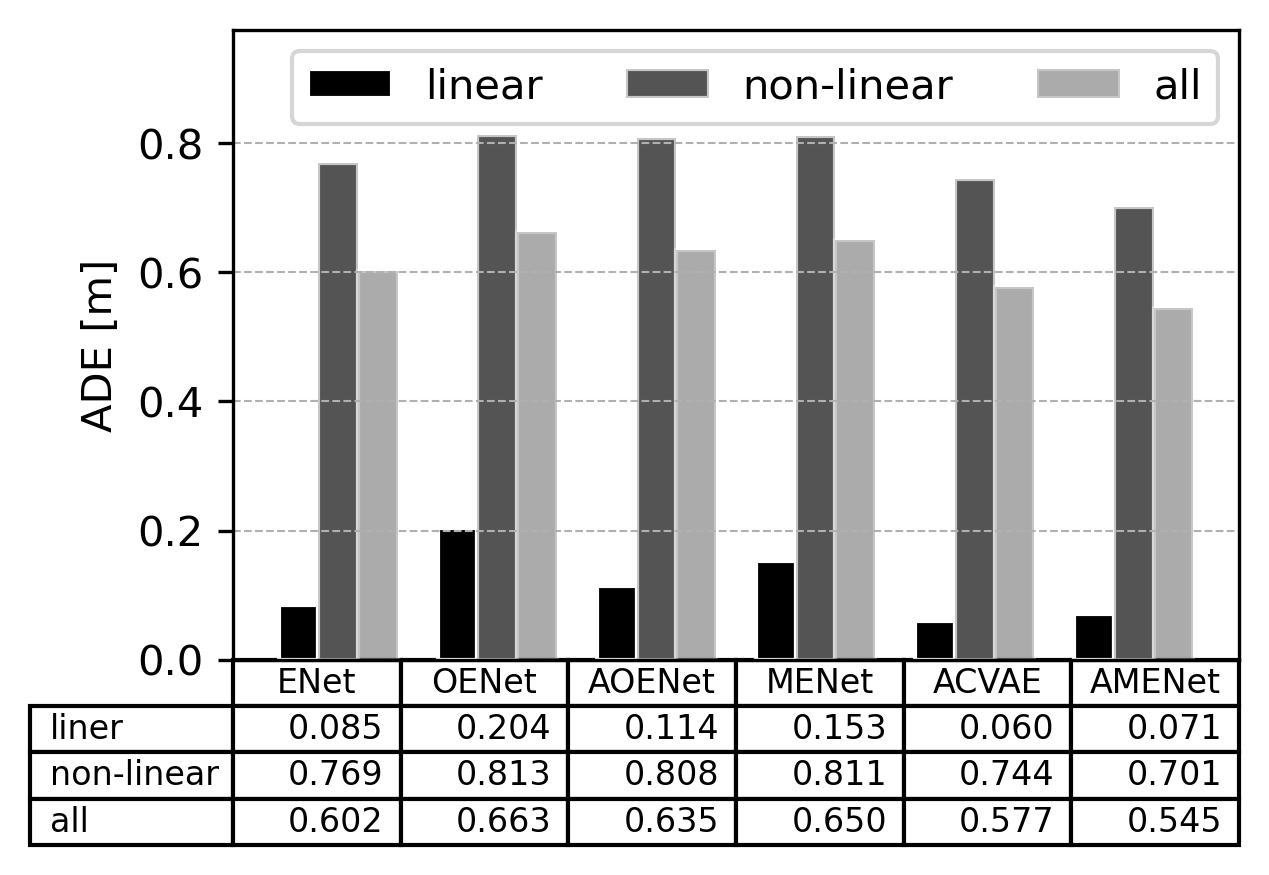}
    \label{subfig:non-linear:ADE}
    \caption{\small ADE}
    \end{subfigure}%
    \begin{subfigure}{0.5\textwidth}
    \includegraphics[trim=0cm 0cm 0cm 0cm, width=1\textwidth]{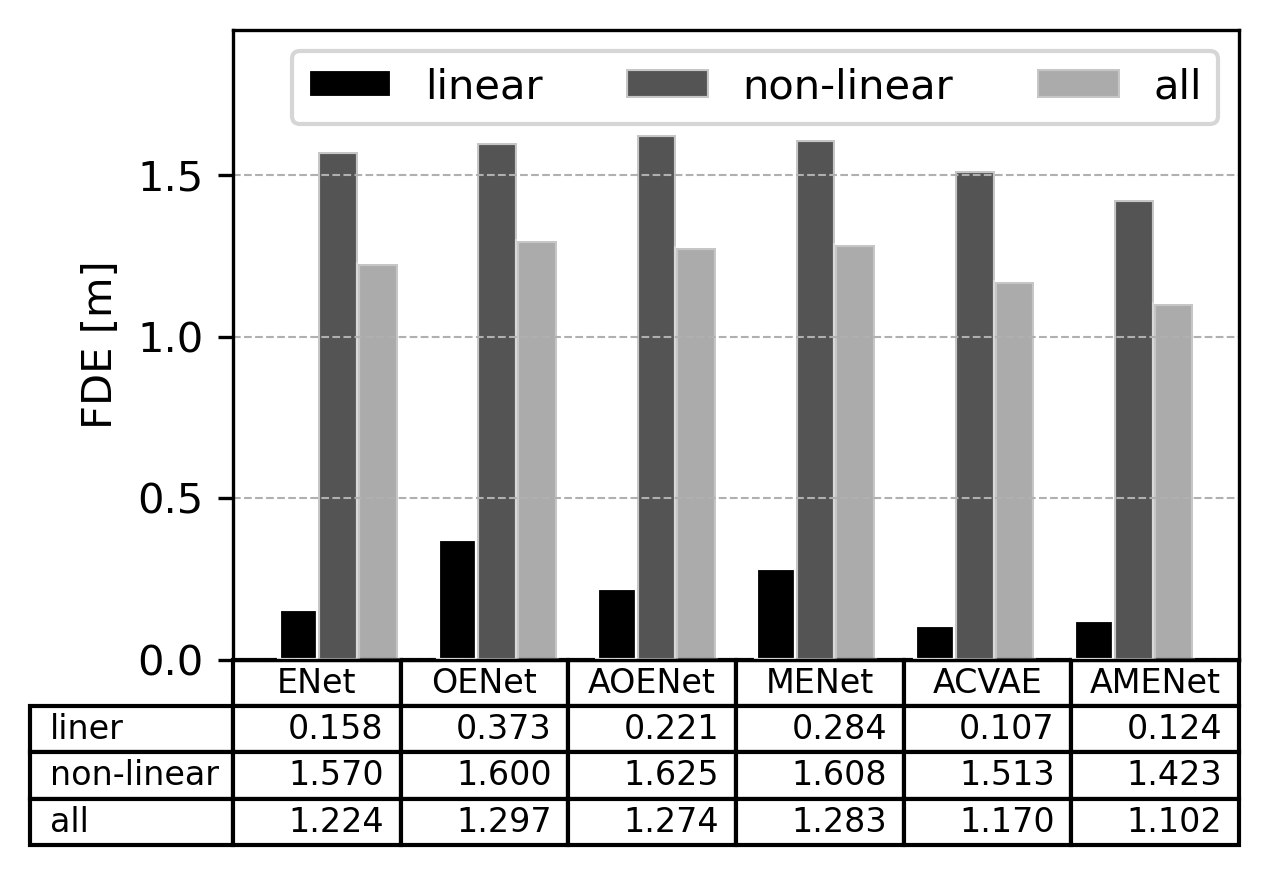}
    \label{subfig:non-linear:FDE}
    \caption{\small FDE}
    \end{subfigure}
    \caption{The prediction errors for linear, non-linear and all trajectories measured by (a) ADE and (b) FDE for all the ablative models, as well as the proposed model AMENet.}
    \label{fig:non-linear}
\end{figure}

The evaluation was decomposed for non-linear and linear trajectories across all of the above models. The linearity of a trajectory not only depends on the continuity of the travel direction, but also on the speed. We use the same scheme as~\cite{gupta2018social} to categorize the linearity of trajectories by a two-degree polynomial fitting. It compares the sum of the squared residuals over the fitting with the least-squares error. A trajectory is categorized as linear if it meets the criteria. Fig.~\ref{fig:non-linear} visualizes the values of (a) ADE and (b) FDE averaged over the six scenes in the offline test set. Across the models, the performance for predicting non-linear trajectories demonstrates a similar pattern compared to predicting all the trajectoires (linear $+$ non-linear) and AMENet outperformed the other models measured by both metrics. Obviously, predicting the linear trajectories is easier than the non-linear ones. In this regard, all the models performed very well (ADE $\leq 0.2$ m and FDE $\leq 0.4$ m), especially the AMENet and ACVAE models. This observation indicates that if there are other agents interacting with each other, the continuity of their motion is likely to be interrupted,~\ie~deviating from the free-flow trajectories~\cite{rinke2017multi}. The model has to adapt to this deviation to achieve a good performance. On the other hand, if there is no such reason to disrupt the linearity of the motion, then the model does not generate deviated trajectories.    

\begin{table}[t!]
\setlength{\tabcolsep}{2.8pt}
\centering
\small
\caption{Evaluation results measured by ADE/FDE on the most-likely prediction for the ablative models and the proposed model AMENet. Best values are highlighted in boldface.}
\begin{tabular}{lllllll}
\hline
Scene    & ENet         & OENet       & AOENet       & MENet       & ACVAE         & AMENet     \\ \hline
B        & 0.532/1.080  & 0.601/1.166 & 0.574/1.144  & 0.576/1.139 & 0.509/1.030   & \textbf{0.486}/\textbf{0.979} \\
C        & 0.241/0.474  & 0.342/0.656 & 0.260/0.509  & 0.294/0.572 & 0.237/0.464   & \textbf{0.226}/\textbf{0.442} \\
D        & 0.681/1.353  & 0.741/1.429 & 0.726/1.437  & 0.725/1.419 & 0.698/1.378   & \textbf{0.659}/\textbf{1.297} \\
G        & 0.876/1.848  & 0.938/1.921 & 0.878/1.819  & 0.941/1.928 & 0.861/1.823   & \textbf{0.797}/\textbf{1.692} \\
H        & 0.598/1.202  & 0.661/1.296 & 0.619/1.244  & 0.657/1.292 & 0.566/1.140   & textbf{0.542}/\textbf{1.094} \\
N       & 0.684/1.387  & 0.695/1.314 & 0.752/1.489  & 0.705/1.346 & 0.595/1.181   & textbf{0.559}/\textbf{1.109} \\
Avg.    & 0.602/1.224  & 0.663/1.297 & 0.635/1.274 & 0.650/1.283  & 0.577/1.170   & \textbf{0.545}/\textbf{1.102} \\ \hline
\end{tabular}
\begin{tabular}{@{}c@{}}
\multicolumn{1}{p{\textwidth}}{B: bookstore3, C: coupa3, D: deathCircle0, G: gates1, H: hyang6, N: nexus6}
\end{tabular}
\label{tb:resultsablativemodels}
\end{table}

\begin{figure}[t!]
    \centering
    \includegraphics[trim=0cm 0cm 0cm 0cm, height=0.88\textheight]{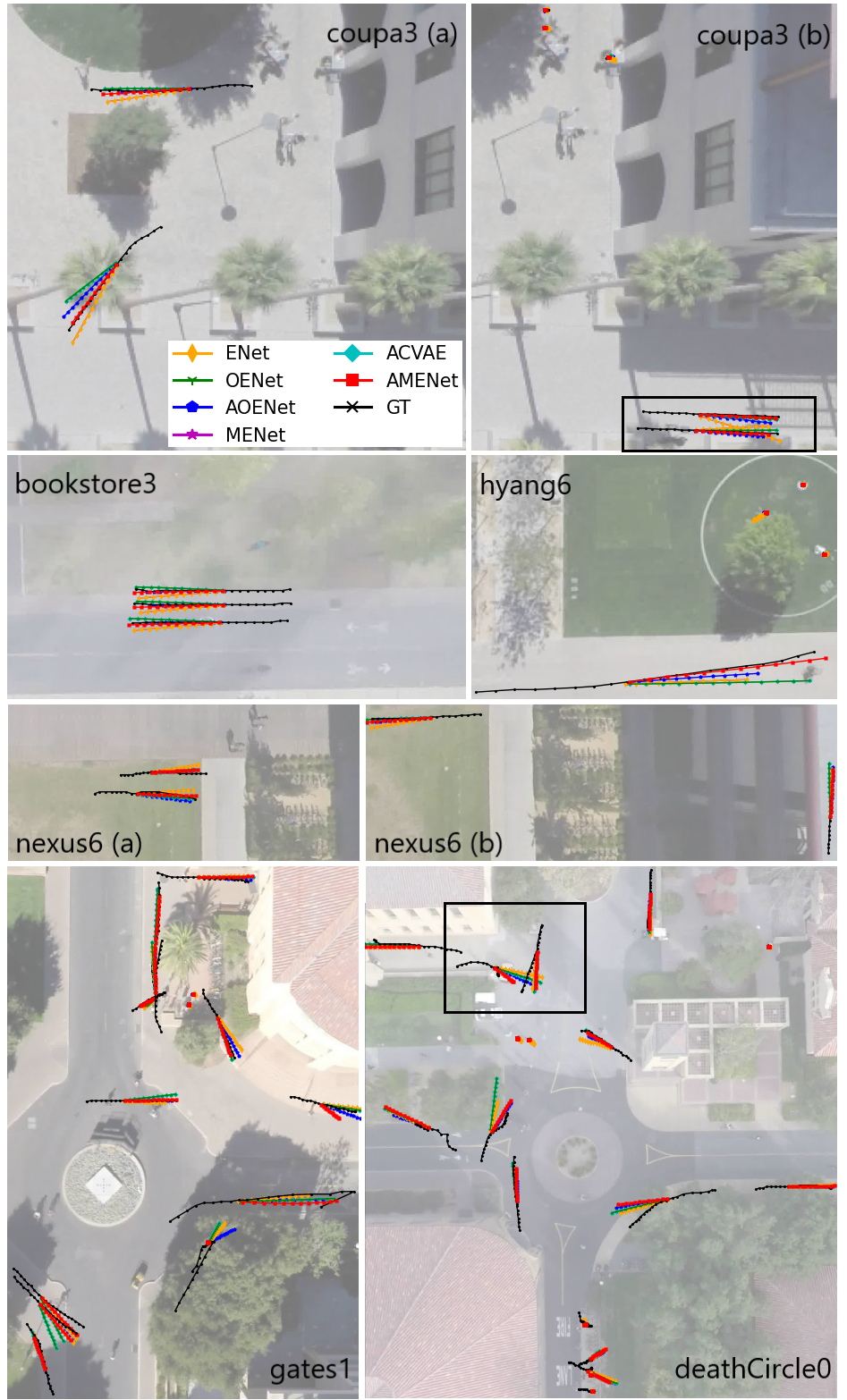}
    \caption{Trajectories predicted by ENet, OENet, AOENet, MENet, ACVAE and AMENet in comparison with the ground truth (GT) trajectories on Trajnet~\cite{sadeghiankosaraju2018trajnet}.}
    \label{fig:abl_qualitative_results_com}
\end{figure}

Fig.~\ref{fig:abl_qualitative_results_com} showcases some qualitative results by the proposed AMENet model in comparison to the ablative models. In general, AMENet generated accurate predictions and outperformed the other models in all the scenes, which is especially visible in coupa3 (a) and bookstore3. 
All the models predicted plausible trajectories for two agents walking in parallel in coupa3 (b) (denoted by the black box), except the baseline model ENet. Without modeling interactions, the ENet model generated two trajectories that intersected with each other. In hyang6 limited performance can be seen by ENet, AOENet and MENet regarding travel speed and OENet and ACVAE regarding destination for the fast-moving agent. In contrast AMENet kept a good prediction. In nexus6 (a) and (b), only two agents were present, where all the models performed well. More agents were involved in the roundabout scenes, in which the prediction task was more challenging. AMENet generated accurate predictions for most of the agents. However, its performance is limited for the agents that changed speed or direction rapidly from their past movement. We notice one interesting scenario of the two agents that walked towards each other in deathCircle0 (denoted by the black box). In reality, when the right agent changed its heading towards the left agent, the left agent had to decelerate strongly to yield the way. Regarding the interaction and compared with the other models, AMENet generated non-conflict trajectories.

\subsection{Trajectory prediction on Benchmark InD}
\label{subsec:InD}
\begin{figure}[bpht!]
    \centering
    \includegraphics[trim=0cm 0cm 0cm 0cm, width=\textwidth]{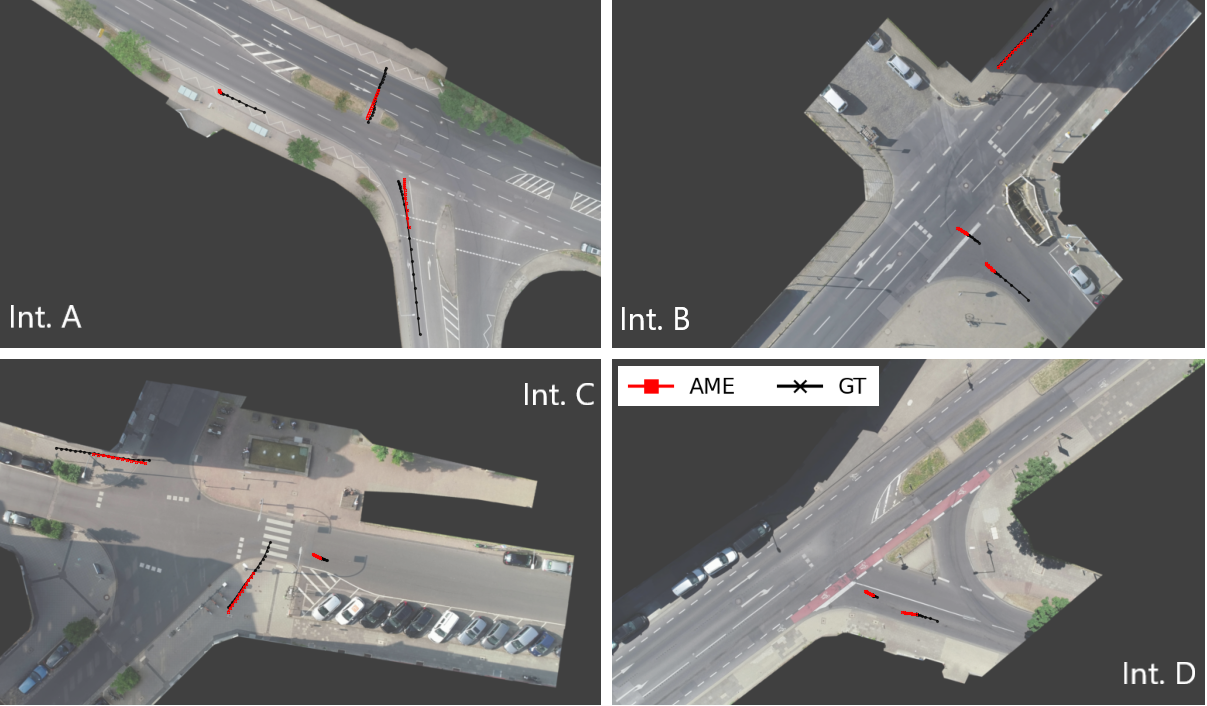}
    \caption{\small{Trajectories predicted by AMENet on the InD benchmark~\cite{inDdataset}.}}
    \label{fig:qualitativeresultsInD}
\end{figure}

To further investigate the generalization performance of the proposed model, we carried out extensive experiments on a newly published large-scale benchmark InD\footnote{\url{https://www.ind-dataset.com/}}.
It consists of 33 datasets and was collected using drones on four very busy intersections (as shown in Fig.~\ref{fig:qualitativeresultsInD}) in Germany in 2019 by Bock et al. \cite{inDdataset}.
Different from Trajnet where most of the environments (\ie~shared spaces  \cite{reid2009dft,robicquet2016learning}) are pedestrian friendly, the intersections in InD are dominated by vehicles. This makes the prediction task more challenging due to the very different travel speed between pedestrians and vehicles, as well as the direct interactions. We follow the same format as the Trajnet benchmark for data processing (Sec.~\ref{subsec:benchmark}). The performance of AMENet is compared with Social LSTM~\cite{alahi2016social} and Social GAN~\cite{gupta2018social}, which are the most relevant ones to our models.
As mentioned above, Social LSTM~\cite{alahi2016social} is the first deep learning method that uses occupancy grid for modeling interactions between agents and Social GAN~\cite{gupta2018social} is the closest deep generative model to ours. It is worth mentioning that we trained and tested all the three models using the same data for fair comparison. 

The performance is analyzed quantitatively and qualitatively. Table~\ref{tb:resultsInD} lists the evaluation results measured by ADE/FDE for all the models in each intersection. AMENet predicted more accurate trajectories measured by all the metrics compared with Social LSTM and Social GAN. Fig.~\ref{fig:qualitativeresultsInD} shows one scenario in each of the four intersections. AMENet predicted the deceleration of the car approaching the intersection from the right arm, the trajectory of the cross walking pedestrian and the slowing down of the pedestrian on the sidewalk in intersection A. It correctly predicted two cars slowly approaching the intersection area in intersection B and D, and the waiting scenario for pedestrian cross walking in intersection C.  
\begin{table}[t!]
\centering
\small
\caption{Quantitative results of AMENet and the comparative modles on InD~\cite{inDdataset} measured by ADE/FDE.  Best values are highlighted in bold face.}
\begin{tabular}{lllllll}
\hline
Model                 & S-LSTM  & S-GAN      & AMENET    & S-LSTM &  S-GAN      & AMENET    \\ \hline
InD           & \multicolumn{3}{c}{@top 10}          & \multicolumn{3}{c}{Most-likely}      \\ \hline
Int. A        & 2.04/4.61    & 2.84/4.91 & \textbf{0.95/1.94} & 2.29/5.33    & 3.02/5.30 & \textbf{1.07/2.22} \\
Int. B        & 1.21/2.99    & 1.47/3.04 & \textbf{0.59/1.29} & 1.28/3.19    & 1.55/3.23 & \textbf{0.65/1.46} \\
Int. C        & 1.66/3.89    & 2.05/4.04 & \textbf{0.74/1.64} & 1.78/4.24    & 2.22/4.45 & \textbf{0.83/1.87} \\
Int. D        & 2.04/4.80    & 2.52/5.15 & \textbf{0.28/0.60} & 2.17/5.11    & 2.71/5.64 & \textbf{0.37/0.80} \\
Avg.          & 1.74/4.07    & 2.22/4.29 & \textbf{0.64/1.37} & 1.88/4.47    & 2.38/4.66 & \textbf{0.73/1.59} \\ \hline
\end{tabular}
\begin{tabular}{@{}c@{}}
\multicolumn{1}{p{\textwidth}}{S-LSTM: Social LSTM~\cite{alahi2016social}, S-GAN: Social GAN~\cite{gupta2018social}}
\end{tabular}
\label{tb:resultsInD}
\end{table}

\subsection{Discussion of the Results}

Based on the extensive studies and results, we discuss the advantages and limitations of the AMENet model proposed in this paper.

AMENet demonstrated superior performance over different benchmarks for trajectory prediction. Firstly, the proposed model was able to achieve the state-of-the-art performance on the Trajnet benchmark challenge, which contains various scenes. Secondly, the results of the ablation studies proved that the information of interactions between agents is beneficial for trajectory prediction. However, the performance highly depends on how such information was leveraged. It was difficult for the occupancy grid, which is only based on the positions of the neighboring users, to extract useful information for interaction modeling, because positions change from one step to the next and from one scene to another. Meanwhile, the speed and orientation information is not considered, which may explain why the occupancy grid performed worse than the dynamic maps in the same settings. Thirdly, as interactions change over time, the self-attention mechanism automatically extracted the salient features in the time axis from the dynamic maps.

However, there are several limitations of the model being uncovered throughout the experiments. First, the resolution of the map was approximated according to the experimental data and the size of the neighboring agents was not yet considered. This may limit the model for dealing with big-sized agents, such as buses or trucks. We leave this to future work. Second, from the qualitative results we notice that the model had limited performance for predicting the behavior of the agents that drastically change direction and speed, which is in general a very challenging task without extra information from the agents, such as body post or eye gaze. Last but not least, in this study, scene context information was not included. The lack of this information may lead to a wrong prediction, \eg~trajectories leading into obstacles or inaccessible areas. Scene context can have a positive effect that a trajectory follows a (curved) path. On the other hand, a strong constraint from the scene context can easily overfit a model for some particular scene layout~\cite{cheng2020mcenet}. Hence, a good mechanism for parsing the scene information is needed to balance the trade-off, especially for a model trained in one scene and applied in another. 

\section{Conclusions}
\label{sec:conclusion}

In this paper, we have presented a generative model called Attentive Maps Encoder Network (AMENet) for multi-path trajectory prediction and made the following contributions.
(1) The model captures the stochastic properties of road users' motion behaviors after a short observation time via the latent space learned by the X-Encoder and Y-Encoder that encode motion and interaction information,  and predicts multiple plausible trajectories. 
(2) We propose a novel concept--attentive dynamic maps--to extract the social effects between agents during interactions. The dynamic maps capture accurate interaction information by encoding the neighboring agents' orientation, travel speed and relative position in relation to the target agent, and the self-attention mechanism enables the model to learn the global dependency of interaction over different steps. 
(3) The model targets heterogeneous agents in mixed traffic in various real-world traffic environments.
The efficacy of the model was validated on the benchmark Trajnet that contains various datasets in different real-world environments and the InD benchmark for different intersections. The model not only achieved state-of-the-art performance, but also won the first place on the leader board for predicting 12 time-step positions of 4.8 seconds. 
Each component of AMENet has been validated via a series of ablation studies.

In future work, we plan to include more information to further improve the prediction accuracy, such as the type and size information of agents and spatial context information. In addition, we will extend our trajectory prediction model for safety analysis, \eg~using the predicted trajectories to calculate time-to-collision \cite{perkins1968traffic} and to detect abnormal trajectories by comparing the anticipated/predicted trajectories with the actual ones.

\section*{Acknowledgements}
This work is supported by the German Research Foundation (DFG) through the Research Training Group SocialCars (GRK 1931).

\bibliography{bibliography}

\pagebreak
\setcounter{equation}{0}
\setcounter{table}{0}
\setcounter{figure}{0} 
\section*{Appendix A: Conditional Variational Auto-Encoder}
\label{appendixA}

\begin{figure}[hbpt!]
\centering
\includegraphics[trim=0in 0in 0in 0in, width=0.3\textwidth]{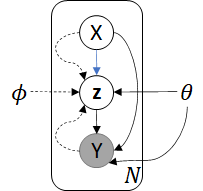}
\caption{Conditional Variational Auto-Encoder graphical model. The generative model $p_\theta\left(\boldsymbol{Y}|\boldsymbol{X},\,\mathbf{z}\right)p_\theta(\mathbf{z})$ is denoted by black solid lines, where $p_\theta(\mathbf{z})$ is the prior of the latent variables. The prior is made independent from the input variables such that $p_\theta(\mathbf{z}|\boldsymbol{X}) = p_\theta(\mathbf{z})$ \citep{kingma2014semi}, denoted by blue solid lines. The variational approximation $q_\phi(\mathbf{z}|\boldsymbol{Y}, \,\boldsymbol{X})$ to the intractable posterior of $p_\theta(\mathbf{z}|\boldsymbol{Y}, \,\boldsymbol{X})$ is denoted by dashed lines.}  
\label{fig:graphicalcvae}
\end{figure}

The Conditional Variational Auto-Encoder (CVAE) model is built upon the variational inference and the learning of directed graphical models \citep{kingma2014auto,rezende2014stochastic,kingma2014semi}, as denoted by Fig.~\ref{fig:graphicalcvae}. CVAE is an extension of the Variational Auto-Encoder (VAE) \citep{kingma2014auto}. For understanding the solution of the CVAE model, it is necessary to revisit the variational inference and the VAE model.

The VAE assumes that the dataset, $\mathbf{X} = \{\boldsymbol{X}_1,\cdots,~\boldsymbol{X}_N\}$, contains $N$ independent and identically distributed samples of some continuous or discrete variable $\boldsymbol{X}$. The dataset can be generated from some unobserved random variables $\mathbf{z}$, the so-called latent variables \citep{kingma2014auto}.  Eq.~\eqref{eq:vae-generation} denotes the integral of the marginal likelihood, where $p_\theta(\mathbf{z})$ is the prior distribution of the latent variables and $\theta$ are the generative parameters.
\begin{equation}
\label{eq:vae-generation}
p_\theta(\mathbf{X})=\int p_\theta(\mathbf{X|\mathbf{z}})p_\theta(\mathbf{z}) d\mathbf{z}.
\end{equation}

However, the equation cannot be solved analytically due to the intractable posterior $p_\theta(\mathbf{z}|\mathbf{X})=p_\theta(\mathbf{X|\mathbf{z}})p_\theta(\mathbf{z})/p_\theta(\mathbf{X})$, or efficiently due to the expensive sampling over a large dataset. To solve the problem, a variational approximation of the true posterior is introduced as the recognition model $q_\phi(\mathbf{z}|\mathbf{X})$, where $\phi$ is the variational parameter. Then Eq.~\eqref{eq:vae-generation} can be rewritten as:
\begin{align}
\label{eq:vae-log}
\log p_\theta(\boldsymbol{X}_1,\cdots,~\boldsymbol{X}_N) &= \sum_{i=1}^{N}\log p_\theta(\boldsymbol{X}_i), \\
\label{eq:vae-kl}
\log p_\theta(\boldsymbol{X}_i) &= D_{KL}(q_\phi(\mathbf{z}|\boldsymbol{X}_i)||p_\theta(\mathbf{z}|\boldsymbol{X}_i)) + \mathcal{L}(\theta, \,\phi; \boldsymbol{X}_i).
\end{align}
It summarizes over the marginal likelihoods of individual data points. The Kullback-Leibler divergence $D_{KL}(q_\phi(\mathbf{z}|\boldsymbol{X}_i)||p_\theta(\mathbf{z}|\boldsymbol{X}_i))$ measures the error of the approximation to the true posterior. Note that the Kullback-Leibler divergence is always non-negative. Hence,
\begin{equation}
\label{eq:vae-lowerbound}
\log p_\theta(\boldsymbol{X}_i) \geq \mathcal{L}(\theta, \,\phi; \boldsymbol{X}_i),
\end{equation}
which is called the (variational) \textit{lower bound} on the marginal likelihood of the data point $i$.
With the variational approximation, Bayes' theorem is applied to solve $\mathcal{L}(\theta, \phi; \boldsymbol{X}_i)$ as follows:
\begin{align}
\begin{split}
 \mathcal{L}(\theta, \phi; \boldsymbol{X_i}) &= \log \mathbb{E}_{q_\phi(\mathbf{z}|\boldsymbol{X}_i)} 
                                              \frac{p_\theta(\boldsymbol{X}_i, \,\mathbf{z})}{q_\phi(\mathbf{z}|\boldsymbol{X}_i)} \\ 
                                            &= -D_{KL}(q_\phi(\mathbf{z}|\boldsymbol{X}_i)||p_\theta(\mathbf{z}))
                                            + \mathbb{E}_{q_\phi(\mathbf{z}|\boldsymbol{X}_i)}
                                            [\log p_\theta(\boldsymbol{X}_i|\mathbf{z})],\label{eq:vae-bayes}
\end{split}
\end{align}
where $-D_{KL}(\,\cdot\,)$ is the negative Kullback-Leibler divergence of the approximate posterior from the prior $p_\theta(\mathbf{z})$ and acts as a regularizer. $\mathbb{E}_{q_\phi(\mathbf{z}|\boldsymbol{X}_i)}(\,\cdot\,)$ is an expected negative reconstruction loss. 
When one optimizes the log likelihood on the left side of Eq.~\eqref{eq:vae-lowerbound}, the Kullback-Leibler divergence and the reconstruction loss are jointly minimized. Hence, the recognition model parameters $\phi$ and the generative parameters $\theta$ can be learned jointly.
The structure of an ``Auto-Encoder'' framework becomes intuitive in Eq.~\eqref{eq:vae-bayes}: the recognition model $q_\phi(\mathbf{z}|\boldsymbol{X_i})$ \textit{encodes} the input into the latent variables and the generative model $\log p_\theta(\boldsymbol{X_i}|\mathbf{z})$ \textit{decodes} the output from the latent variables. 
An analytical solution for the  Kullback-Leibler divergence of two distributions can be found in the Gaussian case \citep{kingma2014auto}. 
The reconstruction error requires sampling, \eg~Monte Carlo sampling.
The lower bound estimation of the VAE model is solved as:
\begin{equation}
\label{eq:vaeestimation}
\mathcal{L}(\theta, \phi; \boldsymbol{X_i}) = \frac{1}{2}\sum_{j=1}^{J}(1 + \mathbf{\log \sigma}_j^2  - \mathbf{\mu}_j^2 - \mathbf{\sigma}_j^2)
                                             + \frac{1}{L}\sum_{l=1}^L (\log p_\theta(\boldsymbol{X_i}|\mathbf{z}_i^l)).
\end{equation}

There is one remaining problem of the sampling process. Neural networks can be used to parameterize the mapping of $\theta$ and $\phi$, which works in the forward pass in the VAE model. However, there is no gradient of the sampling when the neural networks have to be optimized via gradient descent in the backpropogation. To solve this problem, a \textit{re-parameterization} trick \citep{rezende2014stochastic} is introduced to mimic the stochastic property of the latent variables drawn from $g_\phi(\,,\,)$ while maintaining the gradients for backpropogation at the same time. 
\begin{equation}
\label{eq:vae-reparameterization}
\mathbf{z}_i^l = g_\phi(\mathbf{\epsilon}_i^l, \,\boldsymbol{X}_i^l), ~\text{where}~ \mathbf{z}_i^l \sim q_\phi(\mathbf{z}|\boldsymbol{X}_i^l),
\end{equation}
$\mathbf{\epsilon}$ is a noise vector drawn from the distribution of $\mathcal{N}(0, \,\mathbf{I})$. Assuming the posterior approximation $q_\phi(\mathbf{z}|\boldsymbol{X_i}) = \mathcal{N}(\mathbf{\mu}, \,\mathbf{\sigma^2})$, a valid function of $g_\phi(\mathbf{\epsilon}_i^l,\, \boldsymbol{X_i})$ can be formulated as: 
\begin{equation}
\label{eq:vae-reparameterizationfun}
\mathbf{z}_i^l = g_\phi(\mathbf{\epsilon}_i^l, \,\boldsymbol{X_i^l}) = \mathbf{\mu}_i^l + \mathbf{\sigma}_i^l \odot \mathbf{\epsilon}_i^l
\end{equation}


The Conditional VAE (CVAE) is proposed by \cite{sohn2015learning} for structured output prediction. Different from the VAE that reconstructs the input variables with a variational recognition of the posterior, the CVAE generates desirable outputs conditioned on the input variables. To be more specific, given the observation $\boldsymbol{X}$, the latent variables $\mathbf{z}$ are drawn from $p_\theta(\mathbf{z}|\boldsymbol{X})$, the output $\boldsymbol{Y}$ is generated from $p_\theta(\boldsymbol{Y}|\boldsymbol{X},\,\mathbf{z})$. A latent variable $z$ can be drawn multiple times from the distribution $p_\theta(\mathbf{z}|\boldsymbol{X})$. The multi-sampling process of the latent variables $\mathbf{z}$ allows for modeling multiple modes in the conditional distribution of the output variables $\mathbf{Y}$, the so called one-to-many mapping. Then the integral of the conditional probability is defined as follows:
\begin{equation}
    \label{eq:cae} p_\theta(\mathbf{Y}|\boldsymbol{X}) =\int  p_\theta(\boldsymbol{Y}|\boldsymbol{X}, \,\mathbf{z})p_\theta(\mathbf{z}) d\mathbf{z}.
\end{equation}
 Note that in Eq.~\eqref{eq:cae},  the latent variables $\mathbf{z}$ can be made statistically independent of the input variables such that $p_\theta(\mathbf{z}|\boldsymbol{X}) = \sum_{\boldsymbol{X}}p_\theta (\mathbf{z}|\boldsymbol{X})p(\boldsymbol{X}) = p_\theta(\mathbf{z})$ \citep{kingma2014semi}. 

Similar to the VAE, the variational approximation estimation is used to solve Eq.~\eqref{eq:cae} for the intractable posterior. At the datapoint $i$, the log likelihood is denoted by Eq.~\eqref{eq:cvae_app}, where $q_\phi(\mathbf{z}|\boldsymbol{Y}_i,\,\boldsymbol{X}_i)$ is the variational approximation of the true posterior $p_\theta(\mathbf{z}|\boldsymbol{Y}_i,\,\boldsymbol{X}_i)$.
\begin{equation}
\label{eq:cvae_app}
 \log p_\theta(\boldsymbol{Y}_i|\boldsymbol{X}_i) = D_{KL}(q_\phi(\mathbf{z}|\boldsymbol{Y}_i,\,\boldsymbol{X}_i)||p_\theta(\mathbf{z}|\boldsymbol{Y}_i,\,\boldsymbol{X}_i)) + \mathcal{L}(\theta,\,\phi; \boldsymbol{Y}_i, \,\boldsymbol{X}_i).   
\end{equation}
The lower bound of the log likelihood at the datapoint $i$ is solved analogously to the VAE mentioned above:
\begin{align}
 \log p_\theta&(\boldsymbol{Y}_i|\boldsymbol{X}_i) \\
 &\geq \mathcal{L}(\theta, \,\phi; \boldsymbol{Y}_i, \,\boldsymbol{X}_i), \\
 \label{eq:cvae-derivation}
 &= -D_{KL}(q_\phi(\mathbf{z}|\boldsymbol{X}_i, \,\boldsymbol{Y}_i)||p_\theta(\mathbf{z}))
    + \mathbb{E}_{q_\phi(\mathbf{z}|\boldsymbol{X}_i, \,\boldsymbol{Y}_i)}
    [\log p_\theta(\boldsymbol{Y}_i|\boldsymbol{X}_i, \,\mathbf{z})],\\
 &\simeq -D_{KL}(q_\phi(\mathbf{z}|\boldsymbol{X}_i, \,\boldsymbol{Y}_i)||p_\theta(\mathbf{z}))
    + \frac{1}{L}\sum_{l=1}^L (\log p_\theta(\boldsymbol{Y_i}|\boldsymbol{X_i}, \,\mathbf{z}_i^l)).
\end{align}

\section*{Appendix B: The self-attention mechanism}
The self-attention module~\cite{vaswani2017attention} is trained to assign a weight to the dynamic information of each step (denoted as Values ($V$)) based on how much the latent state of the current step (denoted as Query ($Q$)) matches the latent states of the other steps (denote as Key ($K$)). The latent states of $Q$, $K$ and $V$ are computed via three learnable linear transformations with the same input separately:
\begin{equation}
\label{eq:qkv}
\begin{split}
Q =& \pi(\text{Map})W_Q, ~W_Q \in \mathbb{R}^{D\times d_q},\\
K =& \pi(\text{Map})W_K, ~W_K \in \mathbb{R}^{D\times d_k},\\
V =& \pi(\text{Map})W_V, ~W_A \in \mathbb{R}^{D\times d_v},
\end{split}
\end{equation}
where $W_Q, W_K$ and $W_V$ are the trainable parameters and $\pi(\cdot)$ indicates the encoding function of the dynamic maps. $d_q$, $d_k$ and $d_v$ are the respective dimensionalities of the vector $Q$, $K$ and $V$, which are all set to 4 in implementation.
The attention module outputs a weighted sum of the values $V$, where the weight assigned to each value is determined by the dot-product of $Q$ with $K$: 
\begin{equation}
\label{eq:attention}
    \text{Attention}(Q, K, V) = \text{softmax}(\frac{QK^\mathbf{T}}{\sqrt{d_k}})V,
\end{equation}
where $\sqrt{d_k}$ is the scaling factor, $d_k$ is the dimensionality of the vector $K$ and $\mathbf{T}$ stands for transpose.
This operation is also called \emph{scaled dot-product attention}~\cite{vaswani2017attention}.

To improve the performance of the self-attention module, the \emph{multi-head attention}~\cite{vaswani2017attention} strategy is applied as a conventional operation, where a head is an independent scaled dot-product attention module. In this way, the information is attended from different representation subspaces at different positions jointly: 
\begin{align}
\label{eq:multihead}
\begin{split}
    \text{MultiHead}(Q, K, V) &= \text{ConCat}(\text{head}_1,...,\text{head}_h)W_O, \\
    \text{head}_i &= \text{Attention}(QW_{Qi}, KW_{Ki}, VW_{Vi}),
\end{split}
\end{align}
where $W_{Qi}$, $W_{Ki}$, $W_{Vi}\in \mathbb{R}^{D\times d_{qi}}$ are the same linear transformation parameters as in \eqref{eq:qkv} and $W_{O}$ are the linear transformation parameters for aggregating the extracted information from different heads.
Note that $d_{qi} = \frac{d_{q}}{h}$ must be an aliquot part of $d_{q}$. $h$ is the total number of attention heads and we use two heads in the implementation.

\section*{Appendix C: AMENet model architecture and hyper-parameters}
\label{appendixB}

\begin{sidewaysfigure}[hbpt!]
    \centering
    \includegraphics[trim=0in 0in 0in 0in, width=1.2\textwidth]{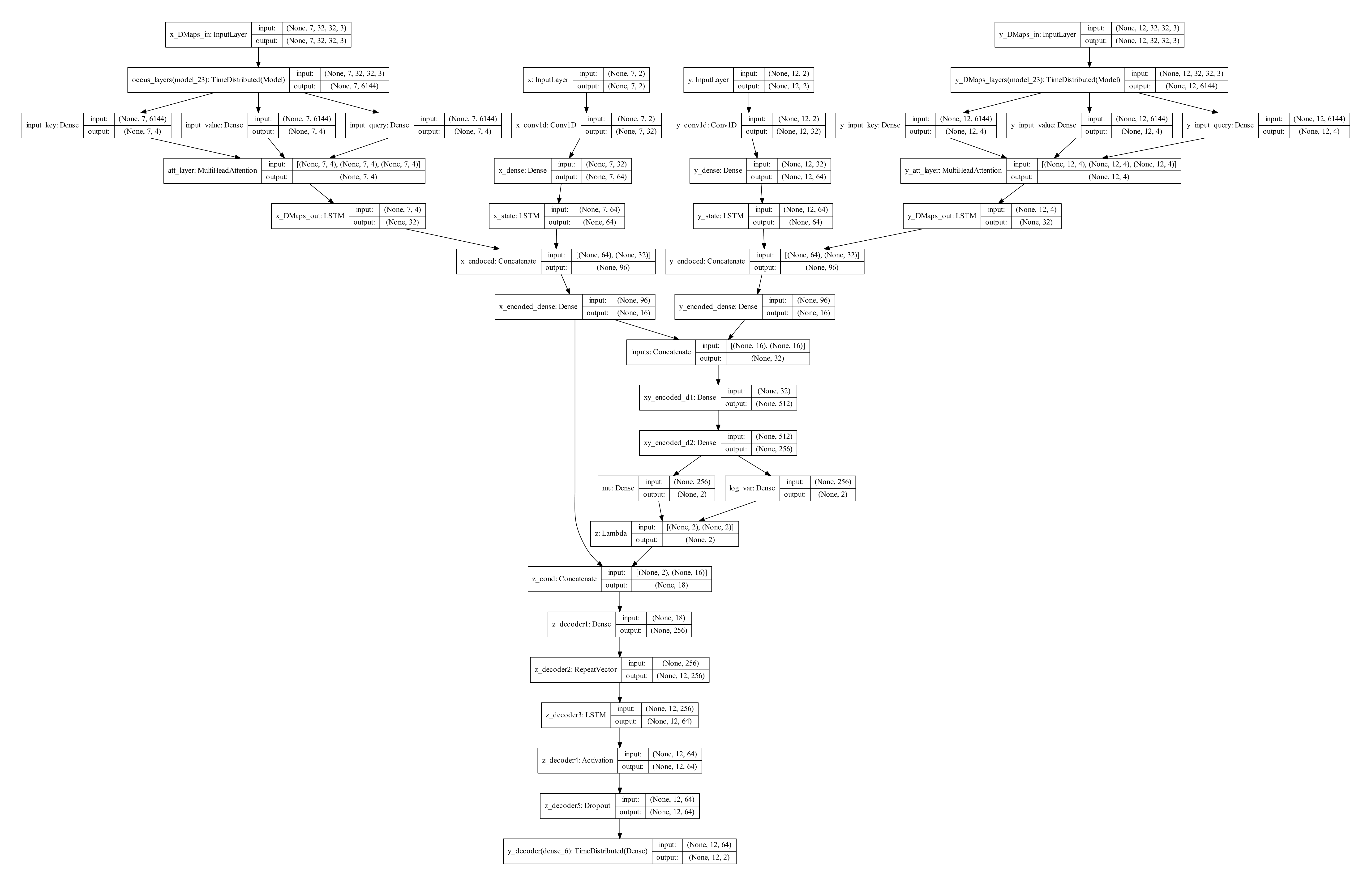}
    \caption{Detailed graph of the AMENet model architecture.}
    \label{fig:AMENET-graph}
\end{sidewaysfigure}

The detailed graph of AMENet is given in Fig~\ref{fig:AMENET-graph}. In the following table we list the most important training hyper-parameters. The detailed settings can be found in our repository at \url{https://github.com/haohao11/AMENet}. 
\begin{table}[hbpt!]
\centering
\small
\caption{Training hyper-parameters for the AMENet model}
\begin{tabular}{lll}
\hline
Name            & Description & Values   \\ \hline
z-dim           & size of the latent variable & 2\\
hidden-size     & size of the LSTM hidden state & 32 \\ 
query           & query dimensionality for the self-attention layer & 4 \\ 
key-value       & key\&value dimensionality for the self-attention layer & 4 \\ 
h               & number of heads & 2\\
beta            & rate of the re-construction loss & 0.70 to 0.85 \\ 
alpha           & rate of the KL-loss & $1-\text{beta}$ \\ 
lr              & learning rate of the Adam optimizer \cite{kingma2014adam} & 0.001 \\ \hline 
\end{tabular}
\label{tb:hyper-parameters}
\end{table}

\end{document}